\documentclass{article}

\usepackage[english]{babel}
\usepackage[square,numbers]{natbib}

\usepackage{PRIMEarxiv}

\usepackage[utf8]{inputenc} 
\usepackage[T1]{fontenc}    
\usepackage{hyperref}       
\usepackage{url}            
\usepackage{booktabs}       
\usepackage{amsfonts}       
\usepackage{nicefrac}       
\usepackage{microtype}      
\usepackage{lipsum}
\usepackage{fancyhdr}       
\usepackage{graphicx}       
\graphicspath{{media/}}     
\usepackage{csquotes}
\usepackage{graphicx}
\usepackage{multicol,caption}

\usepackage{soul}

\usepackage[dvipsnames]{xcolor}

\usepackage{dirtytalk}
\usepackage{hyperref}

\usepackage{amsmath,amssymb} 
\usepackage{subcaption}

\usepackage{multirow}
\usepackage{colortbl}
\usepackage{booktabs, chemformula}

\usepackage{tabularx}
\usepackage{float}
\newcolumntype{Y}{>{\centering\arraybackslash}X}
\usepackage{booktabs}

\title{Spain on Fire: A novel wildfire risk assessment model based on image satellite processing and atmospheric information}

\author{
 Helena Liz-López \\
  Computer Systems Department, \\
  Universidad Polit\'{e}cnica de Madrid, Spain\\
  \texttt{helena.liz@alumnos.upm.es} \\
 \And
Javier Huertas-Tato \\
Computer Systems Department, \\
Universidad Polit\'{e}cnica de Madrid, Spain\\
\texttt{javier.huertas.tato@upm.es} \\
   \And
Jorge Pérez-Aracil  \\
Signal Theory and Communications Department, \\ Universidad de Alcal\'a, spain\\
  \texttt{jorge.perezaracil@uah.es} \\
   \And
Carlos Casanova-Mateo \\
Computer Systems Department, \\
Universidad Polit\'{e}cnica de Madrid, Spain\\
  \texttt{carlos.casanova@upm.es} \\
     \And
Julia Sanz-Justo \\
LATUV, Remote Sensing Laboratory, \\ Universidad de Valladolid, spain\\
  \texttt{julia@latuv.uva.es} \\
     \And
 David Camacho \\
  Computer Systems Department, \\
  Universidad Polit\'{e}cnica de Madrid, Spain\\
  \texttt{david.camacho@upm.es} \\
}

\begin{document}

%

\maketitle

\begin{abstract}

Each year, wildfires destroy larger areas of Spain, threatening numerous ecosystems. Humans cause 90\% of them (negligence or provoked) and the behaviour of individuals is unpredictable. However, atmospheric and environmental variables affect the spread of wildfires, and they can be analysed by using deep learning. In order to mitigate the damage of these events we proposed the novel Wildfire Assessment Model (WAM). Our aim is to anticipate the economic and ecological impact of a wildfire, assisting managers resource allocation and decision making for dangerous regions in Spain, Castilla y León and Andalucía. The WAM uses a residual-style convolutional network architecture to perform regression over atmospheric variables and the greenness index, computing necessary resources, the control and extinction time, and the expected burnt surface area. It is first pre-trained with self-supervision over 100,000 examples of unlabelled data with a masked patch prediction objective and fine-tuned using 311 samples of wildfires. The pretraining allows the model to understand situations, outclassing baselines with a 1,4\%, 3,7\% and 9\% improvement estimating human, heavy and aerial resources; 21\% and 10,2\% in expected extinction and control time; and 18,8\% in expected burnt area. Using the WAM we provide an example assessment map of Castilla y León, visualizing the expected resources over an entire region.

\end{abstract}


\section{Introduction}
\label{section:introduction}

Forests cover 30\% of terrestrial ecosystems, representing a total of 4.06 billion hectares~\cite{mansoor2022elevation} and are home to 80\% of amphibians, 75\% of birds and 68\% of mammals worldwide~\cite{vie2009wildlife}. At the environmental level, in addition to affecting biodiversity (animal and plant), forests are also an important factor in soil transformation, vegetation succession, soil degradation, and air quality, among others~\cite{rodrigues2014modeling}. Wildfires threaten to disturb these ecosystems with increasing frequency and damage, a worrying byproduct of climate change~\cite{senande2022spatial, jones2022global} among many other factors. Fires can also directly affect humans destroying buildings, burning crops, causing the death of animals, or directly have an effect on the health of the population due to the direct action (burns) or indirect action (smoke inhalation) of fire ~\cite{shi2022characterization}.

In recent years the Copernicus Sentinel-3 mission recorded 16000 wildfires throughout the world in August 2018 and 79000 in the same period of 2019, which is a large increase in a single year ~\cite{fernandez2016copernicus}. In Spain, between January and August, the burnt area in 2022 was 247.667 hectares, an area almost five times larger compared to the previous year, 51.571 hectares. In addition, the number of large fires has increased from 16 fires in 2021 to 51 between January and August 2022. Thus, the highest figure since 2012, when 34 large fires occurred, had already been reached \footnote{https://www.epdata.es/datos/incendios-forestales-datos-estadisticas-cifras/267}. Some European countries show an increasing number of wildfires also in the burnt area, as stated by European Forestry Fire Information System (EFFIS). For example, Romania, Italy and France have increased the number of fires and burnt area, showing double the increase in 2022 compared to the annual average between 2006 and 2021 
 \footnote{https://effis.jrc.ec.europa.eu/apps/effis.statistics/estimates}. The growing in severity over the years is concerning. Quality tools are needed to help control and manage resources in order to minimize the damage they cause.

Fire causes that can be aggregated into two groups: natural or anthropogenic. The latter can be divided into two types: \textit{accidental}, due to human negligence; or \textit{provoked}, whether caused by arsonists or pyromaniacs ~\cite{short2022empirical}. In most cases the cause of wildfires is unknown, but if the origin is known human causes account for more than 90\% of the total number of wildfires. This makes fire prediction very difficult, since human behaviour is still unpredictable ~\cite{tiefenbacher2012approaches, menezes2022lightning, pozo2022assessing}. For this reason, an accurate fire prediction model can be considered unfeasible as it would have to rely on individual human behaviour, i.e., when a person is going to commit a reckless act that triggers a fire or when a person is going to decide to start a fire. However, the severity of a fire is tightly related to the existing environmental and vegetation state conditions before its occurrence. Therefore, severity could be estimated observing these conditions ~\cite{cansler2022previous, bonannella2022characterization}.

Atmospheric and environmental variables that influence fire intensity are usually georeferenced variables with a latitude and longitude. Their similarity to images makes this modality analyzable with Computer Vision (CV) techniques such as Convolutional Neural Networks (CNN). These architectures have outstanding ability to identify patterns in data, which enhances the performance of earlier systems based on machine learning models. Due to their strong performance in a variety of tasks, such as computer vision or natural language processing, this advantage has made them a benchmark in deep learning (DL) ~\cite{lecun2015deep}. 

\begin{figure}[!h]
\centering
  \includegraphics[width=0.7\textwidth]{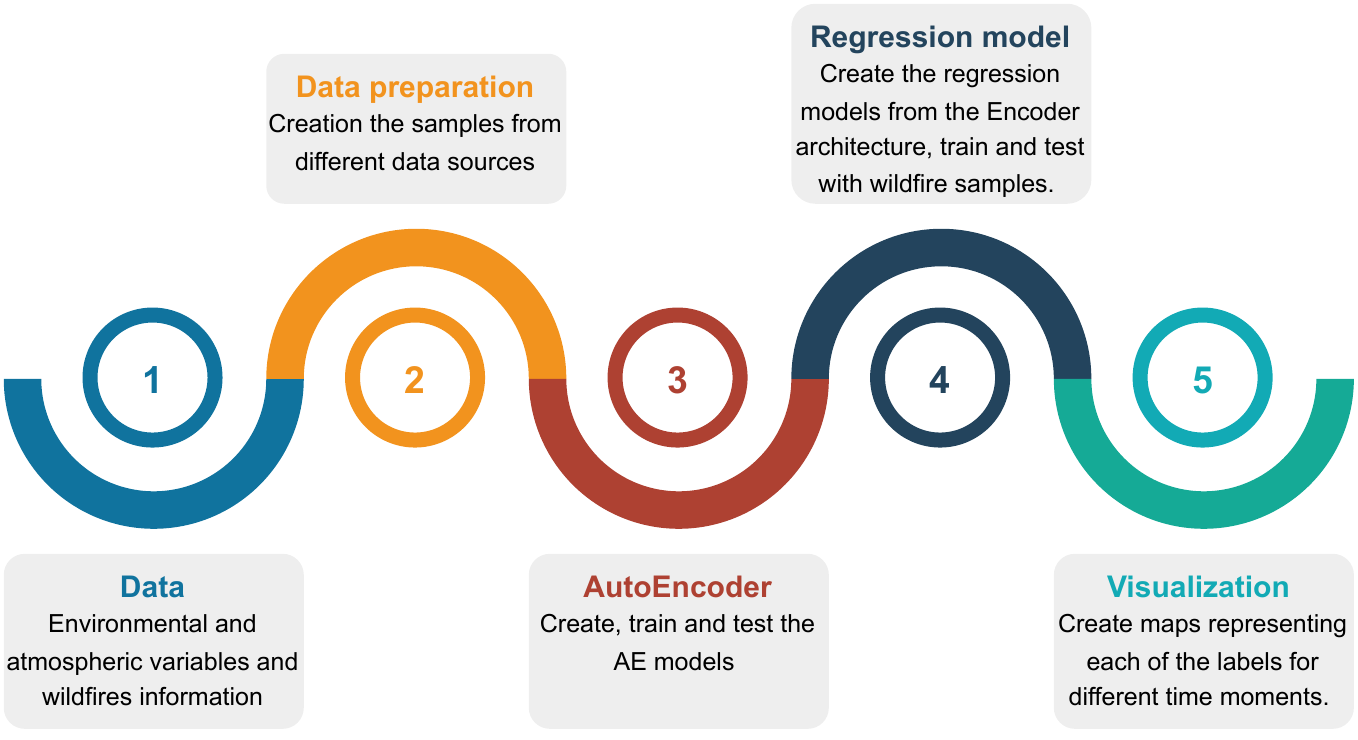}
  \caption{Visual representation of WAM methodology steps.}
  \label{fig:visual_representation_introduction}
\end{figure}

However, most state-of-the-art articles use machine learning techniques, and the few that use DL do not explore options such as CNN. One of the possible reasons for not exploring this type of technique is the lack of data, since most datasets contain a small number of samples. For this reason, we have decided to explore the option of creating a pretrained autoencoder (AE) that is capable of learning the patterns and understanding the atmospheric and environmental variables used. Later, we transfer the encoder to a regression task to predict the variables related to fire management. The article can be divided into five main modules, as shown in Figure \ref{fig:visual_representation_introduction}.

The main contribution of this manuscript is the creation of a regression model based on Deep Learning techniques for wildfire management, called Wildfire Assessment Model (WAM), pretrained with atmospheric and environmental variables of the area and finetuned with a very small data set, 445 samples. However, this is not the only contribution presented in this manuscript:

\begin{itemize}
    \item  The approach used to create the input data is novel and provides more information than traditional approaches.
    \item  WAM model creates a Deep Learning baseline for comparisons for future work in the field.
    \item A new AE model that uses categories instead of the original sample, extracting information on how the meteorological and environmental variables work. 
\end{itemize}

This manuscript has been structured as follows: Section \ref{section:related_work} summarizes the most relevant work in the state of the art, with a special focus on fire severity and burnt area prediction problems; Section \ref{section:data_collection}, describes in detail the atmospheric and environmental variables used, as well as the labels that are intended to be predicted; Section \ref{section:methodology}, describes the proposed methodology for this problem, how we have prepared the samples that are the input to our models, the encoder model and the regression model, the baselines against which we compare ourselves and the final visualization; Section \ref{section:results} presents the experimental results, and Section \ref{section:conclusion_future_work} presents the main conclusions and possible lines of future work. 

\section{Related work}
\label{section:related_work}

Since the 1990s, artificial intelligence has been applied to the science and management of wildfires ~\cite{jain2020review}, as these techniques can contribute to fire detection and management ~\cite{bot2022systematic}. The features used in this field, such as atmospheric, vegetation, or topological variables, can be analysed using DL techniques. These techniques can extract information from samples faster and more efficiently than humans for specific tasks ~\cite{dhankar2021systematic}, and their effectiveness can be observed in other problems related to ecology ~\cite{fister2022accurate,salcedo2022analysis,cornejo4231494machine,torre2019estimation} and the environment ~\cite{sahu2022self,xiang2022impact, mukherjee2022intelligent}.

Within the field of wildfire management, we can find different types of tasks before, during and after the event. At the level of fire prevention, we can find articles on lightning prediction to try to prevent fires at those points ~\cite{nadeem2019mesoscale} or to predict which areas have favourable fire conditions ~\cite{tonini2020machine}. During the active period of wildfires, these techniques can be used to predict which area will burn based on the evolution  ~\cite{hodges2019wildland}, the spread and growth of the fire ~\cite{radke2019firecast}. Finally, machine learning techniques can be applied to post-fire management, such as managing area regeneration ~\cite{guz2021long}, the socioeconomic effects of the incident ~\cite{farasin2020double} or the long-term effects on the substrate, such as erosion ~\cite{buckland2019using}, and how air quality and particulate matter will evolve ~\cite{zou2019machine,michael2021forecasting}.

\subsection{Wildfire prediction systems}
\label{subsec_related_work_prediction}
As we have introduced above, one of the most explored lines of research in this field is the prediction of wildfires. These systems try to predict whether a fire will be caused by a series of atmospheric, topological, vegetation, and human conditions in a given area. ~\citet{campos2021assessing} attempts to predict the probability of fires in tropical dry forests in an area of Costa Rica (Guanacaste Conservation Area) between 1997 and 2020 using a generalized least squares model (GLS), using different variables: slope, accessibility, fire line density and burnt area. ~\citet{xu2021temperature} studied the influence of climate warming on the frequency of wildfires in Changsha (China) between 2011 and 2017, for which they created a dataset consisting of 20622 fire incidents, using atmospheric variables: wind, precipitation, and daily maximum and minimum temperature. However, the previous variables have been combined and finally three variables have been used: every day's fire frequency, daily maximum temperature, and daily minimum temperature. In this article, they used three different models: random forest, SVM, and polynomial, to compare the performance of the different models, they used the MSE metric and $R^{2}$, which showed that the best model was the random forest followed by the polynomial regression model, both using the minimum daily temperature variable. Although the results are promising, the variables used in this study are too limited.

~\citet{janiec2020comparison} studied the risk of fire among six different classes, i.e. a multiclass problem in Yakutia (Russia) between 2001 and 2018. They used different environmental, vegetation, topographic and human variables and conducted two studies, the first at the level of the whole region (Yakutia) and the other of a more specific area (Nyurbinsky), for each of which two different models, random forest, and maximum entropy, each of which performed better in one region, Yakutia and Nyurbinsky, respectively. Another example of fire prediction is ~\citet{milanovic2020forest}, that analyse the probability of fires in eastern Serbia between 2001 and 2018, divided in five different categories (very low, low, moderate, high, and very high), although unlike other articles, they do not specify the number of samples used, each sample is made up of a total of 17 variables that we can classify into 4 different groups: vegetation, anthropogenic, topographic, and climatic. The aim of the article is to predict the risk of fire with two different models, random forest and logistic regression, and finally to show the probabilities in a map format. ~\citet{kang2020developing} show a system to predict the probability of fires in southern Korea between 2014 and 2019. To do so, they classified fires into four risk classes using different meteorological, geographic and fuel data. To analyse these data, they used an ensemble composed of two catboost models where the input was different, the first model used all variables while the second one only used the atmospheric variables. ~\citet{michael2021forecasting} have developed a fire probability mapping system for Greece during 2007, dividing fires into five different categories. The machine learning techniques used were logistic regression, random forest, and XGBoost. Regarding the variables used for this problem they did not rely only on static climatic and topological variables of the study area, but also included dynamic vegetation variables such as NVDI. The best performing model for this problem was XGBoost. ~\citet{pham2020performance} attempt to predict fire risk in Vietnam's e Pu Mat National Park, using a total of 57 wildfire samples, having each one of them nine variables (basically anthropological, vegetation, meteorological and topological). To analyse these data, they used different algorithms, Bayes Network, Naive Bayes, decision tree, and Multivariate Logistic Regression and tried to generate a map from the predicted classes (divided into three different classes). The best performing model was the Bayes network with an AUC = 0.96, followed by the decision tree with an AUC = 0.94.

We can observe how many authors use machine learning techniques such as random forest or Logistic Regression among others to predict wildfires. This is striking considering the great progress of Deep Learning and the fact that the variables used are usually geo-referenced, which also allows them to be analysed with computer vision techniques such as convolutional neural networks. However, we can find some examples that use deep learning. For example, ~\citet{choi2020fire} analyse the risk of fires in Korea between 2011 and 2017. The dataset consists of 201.082 samples composed of 107 different variables that cause and prevent wildfires. To improve current prediction systems, the article uses three different systems: logistic regression, artificial neural networks, and fire risk indexing, which are combined to calculate the final fire risk assessment. ~\citet{mohajane2021application} developed five different ensemble-based models to predict wildfire risk: random forest, support vector machine, multilayer perceptron, which is a neural network, classification and regression tree, and logistic regression. The study area was in the north of Morocco between 2005 and 2015, of which they have a total of 510 wildfire samples, containing a total of 10 variables (topographic, socioeconomic and meteorological). 

Regarding most recent works, it can be seen that Deep Learning techniques are barely used in forest fire studies. This could be explained because of the limited number of samples used in the different works analysed; for that reason we have decided to explore the use of these techniques in a problem of wildfire risk assessment overcoming the limitation set by the number of samples. Another limitation is that the authors try to predict whether a fire will occur or not at a specific point or region, which is extremely difficult since in many cases only meteorological and topological variables are taken into account, which would allow only the prediction of natural fires. Even if anthropological variables are taken into account, such as distance to nearest towns or the existence of routes such as roads, it does not come close to reality, since as mentioned in Section \ref{section:introduction} there are two main reasons for human causes: accidental and provoked, which can be caused by arsonists and pyromaniacs, and human behaviour is impossible to predict. For all these reasons, we decided against creating a wildfire prediction system, due to the number of variables involved and the difficulty of predicting human behaviour.

\subsection{Use of convolutional neural networks in active fire detection}

Other works are focused on the rapid detection of active fires, where we can observe a greater presence of neural networks, in particular convolutional neural networks. ~\citet{muhammad2018early} have developed a system for the detection of active fires in forests, houses, and vehicles, for which they have created a dataset with 68,457 samples and were processed by a convolutional network composed of five convolutional layers and three dense layers, reaching an accuracy greater than 0.94. ~\citet{park2019dependable} combines multiple AI frameworks, including a CNN, a deep neural network, and adaptive fuzzy logic for fire detection. The aim of the system is early detection of active fires in order to be able to carry out a rapid response and minimise damage; the system is focused on finding a solution to the transmission of information, which is the main bottleneck of these systems. ~\citet{zhang2016deep} show another example of early detection of active fires in images. For this purpose, they divided each image into 16x16 patches and tried to detect on which patches flames appeared. For this, they use a dataset composed of 25 videos, including a total of 21 positive and 4 negative sequences. They use two different architectures, the first one a standard CNN composed of three convolutional layers and two dense layers; and the second architecture, based on the previous one, which is trained with local patches that do not contain global information of the sample. Although these models are interesting and use similar techniques to ours, they are only useful when the fires are active, while our objective is to anticipate them, in order to manage resources efficiently and minimise the damage caused by the fire. These models have two main limitations. Firstly, they require security cameras to detect fires, which limits their application. These systems are suitable for urban areas or nearby areas, however, they are unfeasible in forested areas, as it would be difficult to monitor the entire forest area.  Secondly, these systems are designed to detect signs of active fires, such as flames or smoke, which allows a quick reaction to such an event but not prevention or pre-fire management, as is our objective in this work.

\subsection{Machine learning for wildfire management}
Wildfire management can be carried out in different ways, either during the event or before it. One of the options to try to minimise the damage caused by these events is to predict how the fire will spread. For example, ~\citet{zohdi2020machine} shows a system to predict the evolution of active fires based on the trajectories of hot particles, wind, updraughts and the topology of the surrounding combustible material. The authors created a simulation combined with machine learning algorithms to try to predict the dispersion of the fire and the possible generation of new ignition sites. The results obtained by the system are plotted in three dimensions to show in which direction the fire will spread. ~\citet{julian2018autonomous} develop a deep reinforcement learning system using different decentralized controllers that accommodate the high dimensionality and uncertainty inherent in wildfires. For these, they tried two different approaches: the first by controlling aircraft decisions individually, and the second allow the aircraft to collaborate between them on a map and keep track of visited locations. To do this, they developed a stochastic system to simulate fire conditions and see how the fire will progress over time. The architecture presented consists of two different networks, one to process the images based on a typical convolutional neural network and another network composed of dense layers for the continuous inputs. Both outputs are merged and the final result is given by the input of 3 dense layers. ~\citet{mccarthy2021deep} developed a system to predict the evolution of active wildfires, predicting the direction and area that will burn over time. The main problem with this approach, as they explain, is that they are difficult to predict due to the complexity of the problem. Consequently, they proposed a convolutional neural network based on the U-net and used geostationary multispectral images and environmental variables such as vegetation and terrain data. The results are displayed on a map, where the space that the fire will occupy every half hour is marked with different coloured lines. ~\citet{hodges2019wildland} created a series of samples from atmospheric variables, where each channel represents a different variable for the study area, corresponding to satellite measurements. They used 13 different variables related to vegetation, meteorology, topography and other variables, such as the initial burn map or the fuel model. They used the convolutional inverse graphics networks (DCIGN) model to predict the burn map, which is composed of two convolutional blocks followed by a reshape layer (flatten layer), a dense layer and finally a combination of a reshape layer and a convolutional layer that will generate the burn map, 6 and 24 hours after the beginning of the wildfire. Other works try to predict the final size of the affected area between different classes at the time a fire starts. For example, ~\citet{coffield2019machine} tries to predict the final size of the fire between 3 different classes based on meteorological, topographical and vegetation variables. They applied different machine learning algorithms, such as decision tree, random forest, k-nearest neighbours, gradient boosting, and multilayer perceptron. The problem with the latter article is that once the fire starts, the specialized crews can already see how big the fire is going to be.

Some articles tried to predict how susceptible a particular area will be to wildfires. This approach, unlike the previous, allows to better manage resources, try to prevent and minimise the damage caused by fires. ~\citet{boulanger2018model} develops a consensus model based on five different machine learning models: generalized linear model, random forest, gradient-boosted, regression trees, and multivariate adaptive regression splines to project future burning rates in boreal Canada. The data used to design this system were taken between 1980 and 2000 and only fires where the number of hectares burnt was greater than 200 were taken into account. In total, 12228 samples were acquired using a series of climatic, vegetation, land cover, and topographic variables. Future climate projections were obtained from three climate models and three scenarios of anthropogenic climate forcing for three 30-year periods. ~\citet{pais2021deep} have developed a system to predict how susceptible an area is to fires based on topology vegetation variables between 2013 and 2015. They combined the information from all these variables to generate an image that can be analysed using a convolutional neural network, composed of three convolutional blocks of 1, 2 and 3 convolutional layers each and a dense layer. The system returns the ignition probability of each image and a GradCAM-based visualization that allows a better understanding of how the CNN has arrived at the final result. On the other hand, ~\citet{zhang2021deep} generate a global map of the probability of ignition, i.e. how susceptible each point is to fire using vegetation and environmental variables, such as wind, maximum and minimum temperature, humidity or soil moisture, among others. To generate the global fire susceptibility maps they tested four different techniques, two convolutional neural networks, and two MLP models. The first CNN uses two-dimensional convolutional layers and the second one uses one-dimensional convolutional layers; the same happens with the MLP models; the first one uses two-dimensional images flattened with a flatten layer as input, and the second one uses a one-dimensional array as input. As we can see, several authors generate maps from the results obtained by the different systems; this allows us to visualise the probability or risk of the fire, the susceptibility of the area to these events, the burnt area, and the spread of active or in a future wildfire. For example,  ~\citet{al2021wildland} have developed a map where the wildfire susceptibility of each pixel in the study area can be seen. To do this, they have created a system of six classes that represent different degrees of susceptibility (very low, low, moderate, high, and very high); additionally, it is observed that most wildfires have occurred in areas with high susceptibility. Other authors, as ~\citet{zhang2021deep} generate maps on which measure the probability of ignition with a colour scale. 

As with the work shown in Section \ref{subsec_related_work_prediction}, most fire management systems use machine learning algorithms without exploring other options such as deep learning, which has worked very well on other problems. Other work focuses on measuring the susceptibility of a fire to occur, which is another way of trying to predict fires, which as we have already explained is very complex as human behaviour is involved. The models that are really focused on management are those that try to predict the evolution of fires, these works try to predict the area that will burn throughout the wildfires, i.e. they focus on a single label, whereas we have tried to predict six labels related to fire management, including the resources needed and the time needed to control and extinguish fires.

In our work, we have tried to overcome several of the limitations present in the state of the art. First, as in most previous studies, we have an extremely small labelled dataset (597 wildfire data records), so we decided to apply few-shot learning techniques by creating an autoencoder that was able to learn the patterns of the different features of the samples and applying this knowledge to a regression task to predict the resources that would be needed in case of a wildfire. Finally, to facilitate its use by the final users, we have generated prediction maps that shows the resources needed in each area.

\section{Data collection}
\label{section:data_collection}
The objective of this paper, as explained in Section \ref{section:introduction}, is to develop a regression model (WAM) capable of predicting the resources, the time needed to control and extinguish wildfires, and the area that will burn during the event, to help manage resources and reduce economic and environmental losses.

For the purpose of this work, we have used three different sources of information: 1) \textit{wildfire information}, this information will be used to create the labels, including information on the locations where fires have occurred, including the resources needed to extinguish them, the time it took to control and extinguish the fire and the total area burned; 2) \textit{atmospheric variables}, related to fire spread such as 10 metre U wind component or total column ozone; 3) \textit{greenness index}, an indicator of vegetation health is another variable closely related to forest fire management. The last two will be combined using early fusion techniques to create the samples of the different fires (variable X) and the information on fire management will be the labels of these samples.

\subsection{Wildfires in Spain}
\label{subsec:dc_wildfires}
\begin{figure}[!h]
\centering
  \includegraphics[width=3.50in, scale=1]{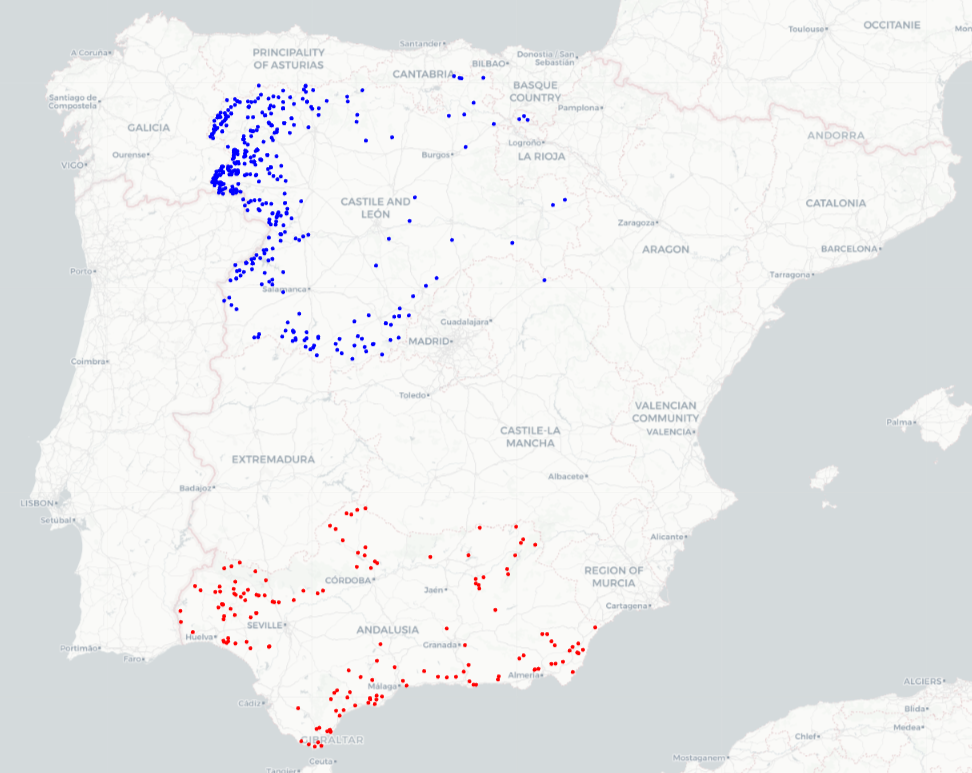}
  \caption{Map of Spain showing the locations of all wildfires recorded between 2001 and 2012. Wildfires in Castilla y León are highlighted in blue and those in Andalucía in red.}
  \label{fig:map_cyl_and}
\end{figure}

\begin{figure}[!h]
\centering
\begin{subfigure}{0.40\textwidth}
\centering
\includegraphics[width = \textwidth]{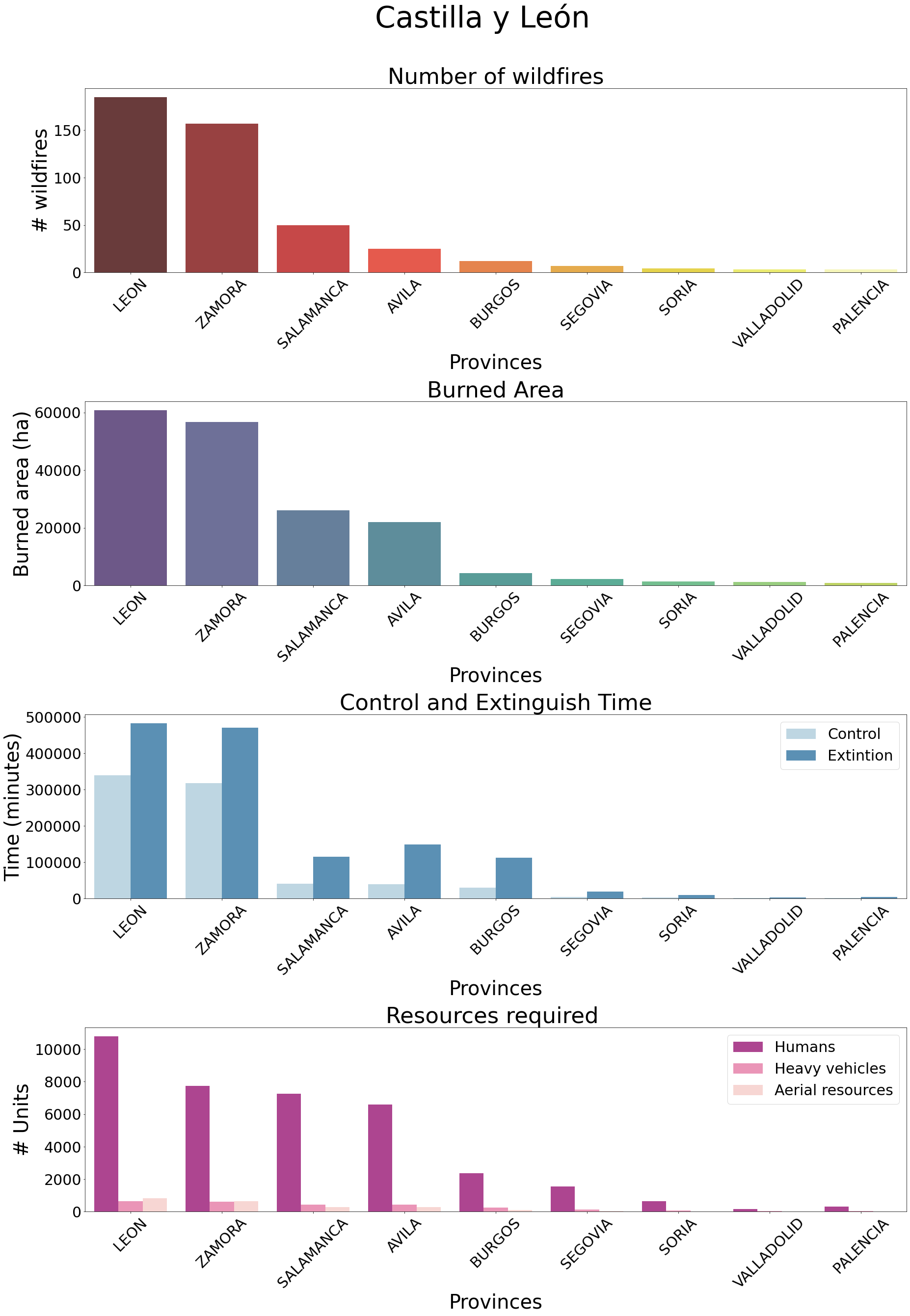}
\caption{Castilla y León}
\label{fig:cyl_plots}
\end{subfigure}
\begin{subfigure}{0.40\textwidth}
\centering
\includegraphics[width = \textwidth]{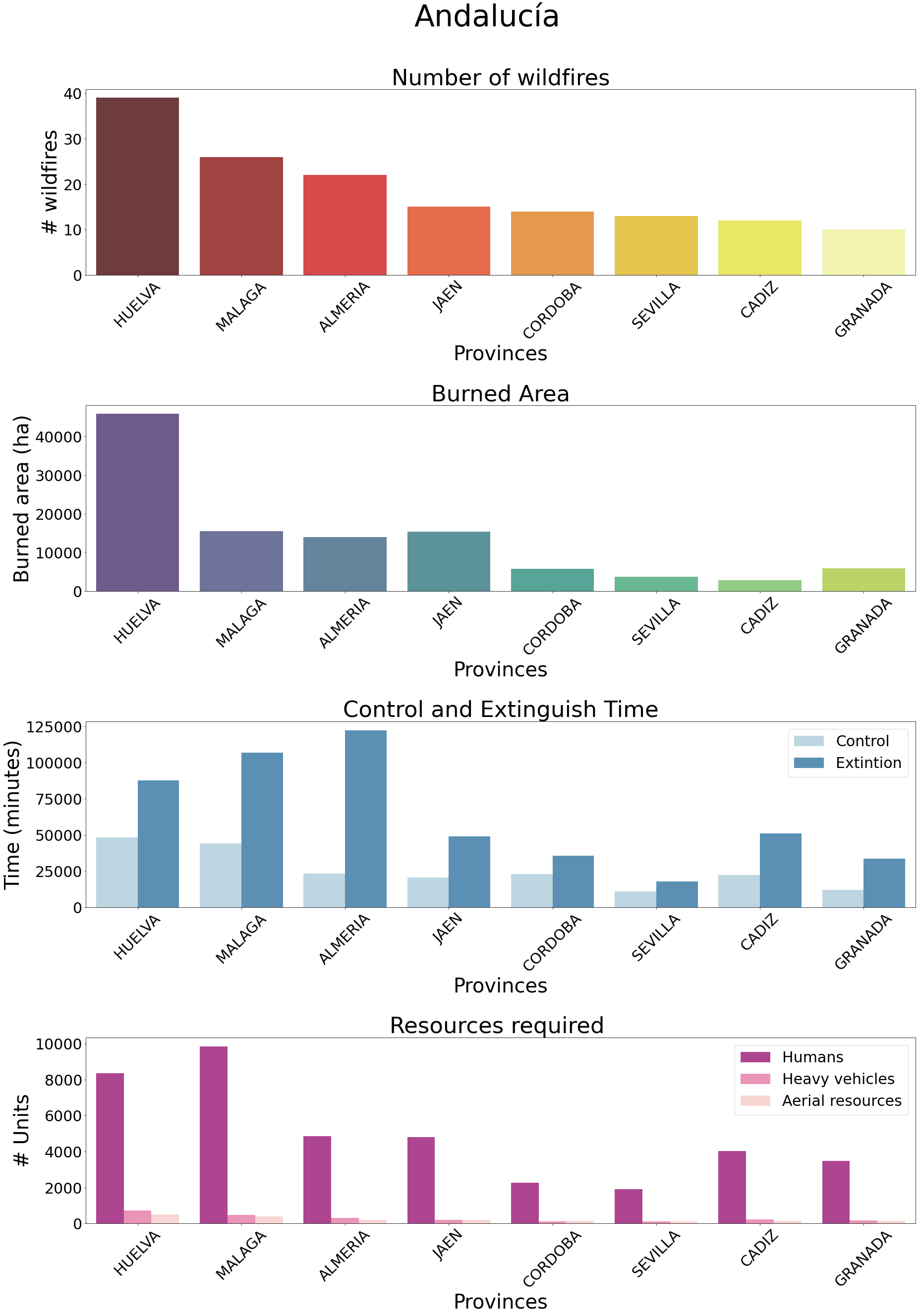}
\caption{Andalucía}
\label{fig:andalucia_plots}
\end{subfigure}
\caption{Visual representation of labels grouped both by province and autonomous region.}
\label{fig:plots}
\end{figure}

In this work, we used a dataset composed of 597 wildfire data records taken from two Spanish autonomous regions: Castilla y León (446) and Andalucía (151), see Figure \ref{fig:map_cyl_and}. Each wildfire data record contains the coordinates, date, and relevant information about the wildfire. The coordinates of Castilla y León are latitudes between 40 and 43.3 and longitudes between -7.2 and -1.8; and the coordinates of Andalucía are latitudes between 35.5 and 38.5 and longitudes between -7.6 and 0. We can see that both regions are very close to each other, but there are significant differences; although they are two autonomous regions in Spain, they have totally different weather. Regarding the annual temperature, there are differences of more than 5 degrees between the both regions, the temperature of Castilla y León is colder. The annual rain, Andalucía has a greater diversity of environments, as it has arid areas (east of Andalucía) and more damp areas (west), while Castilla y León is more homogeneous, with a lower average, except in mountainous areas. Finally, with regard to radiation and insolation, there are also differences between the two, in Andalucía is higher than Castilla y León. Therefore, we can see that the conditions in both regions are very different and that wildfires management must be adapted to them. In Castilla y León the wildfires tend to be concentrated in two sub-regions, Zamora and León, and in others, such as Valladolid, Soria or Palencia, the number of wildfires is insignificant, which means that the fires are concentrated in the west. In Andalucía, wildfires are more distributed throughout the autonomous region, Figure \ref{fig:andalucia_plots}, with fewer events in the centre. Huelva, Málaga, Jaén and Almería are the most affected areas (number of events and burnt area). Although note that forest fires are more spread in Andalucía than in Castilla y León.

As mentioned above, each wildfire data record contains relevant information on fire management, such as extinction time or burnt area. This information is what we will use in our article as labels, each one is explained below. 

\begin{enumerate}
    \item \textbf{Burnt area} (metres): total area affected by wildfire. 
    \item \textbf{Control time} (min): time required to enter the control phase, that is when the fire conditions have changed enough to prevent its propagation.
    \item \textbf{Extinction time} (min): time until the fire is extinguished, that is, when there are no active hotspots and the technicians verify that there is no possibility of reignite.
    \item \textbf{Human resources} (units): number of people involved in extinguishing the fire.
    \item \textbf{Aerial resources} (units): number of aerial vehicles involved in the extinguishing of the fire.
    \item \textbf{Heavy resources} (units): number of heavy vehicles involved in the extinguishing of the fire.

\end{enumerate}

Regarding the Andaluca region, Figure \ref{fig:andalucia_plots}, can be seen some apparent unbalances between the burnt area and the time and resources needed to its extinction, as in the case of the subregion of Almera. On the other hand, in the case of Cordoba, Sevilla, and Cadiz subregions, we see an apparent lower burnt area compared to the number of fires. Another significant difference is that although both regions, Andalucía and Castilla y León, show a similar burnt area, the time and resources needed to extinguish wildfires are higher in Castilla y León. This could be related not only to the topography but also to the previous history of the burnt area (temperature, moisture, vegetation dryness, etc.).

 \subsection{Atmospheric variables}
\label{subsec:dc_wildfires}

As mentioned above, the second source of information we have used is a set of atmospheric variables related to the spread of forest fires. As these data sources are geo-referenced, i.e. they can be arranged in two-dimensional matrices, as if they were images. These variables use a decimal coordinate system, and the distance between the measured latitudes is 111 km. These data were measured daily in our study period, depending on the variables, they were measured every hour or at different times of the day (always at the same times). These variables are explained as follows ~\cite{hersbach2020era5}:

\begin{itemize}
    \item \textit{10 metre U wind component}: is the horizontal speed of air moving toward the east at a height of 10 metres above the Earth's surface. This variable is positive if the wind comes from the west ~\cite{baars2021californian}.
    
    \item \textit{10 metre V wind component}: it is the vertical speed of air moving north at a height of 10 metres above the Earth's surface. 10 metre V wind component is positive if the wind comes from the south ~\cite{baars2021californian}.
    
    \item \textit{2 metre dewpoint temperature} (K, kelvin): is the temperature that would have to be cooled for saturation to occur at 2 metres above the Earth's surface. It is a measure that combines humidity, temperature, and pressure. This parameter is calculated as an interpolation between the lowest level of the model and the Earth's surface, considering the atmospheric conditions ~\cite{haiden2018addressing}.
    
    \item \textit{Surface net solar radiation}(J/m$^{2}$): is the low-wave solar radiation incident on the Earth's surface, direct and indirect, less the reflected radiation. It is the radiation that crosses a horizontal plane to the Sun's direction ~\cite{alados2003relationship}.
    
    \item \textit{Surface net thermal radiation} (J/m$^{2}$):is the radiation emitted by the atmosphere, clouds, and the Earth's surface. It is the difference between downward and upward thermal radiation at the Earth's surface. The downward thermal radiation is the radiation reaching the surface emitted by clouds and the atmosphere; and the upward radiation corresponds to the radiation emitted by the surface and the downward radiation reflected by the surface ~\cite{hogan2015radiation}. 
    
    \item \textit{Surface solar radiation downwards} (J/m$^{2}$): describes the amount of shortwave solar radiation that reaches the earth's surface on a horizontal plane. It represents the radiation incident on the Earth's surface, i.e. that which is not reflected by clouds, suspended particles, etc. ~\cite{yuan2021global}.
    
    \item \textit{Surface thermal radiation downwards} (J/m$^{2}$): is a type of thermal radiation that describes the thermal radiation reaching the earth's surface emitted by the atmosphere and clouds ~\cite{hogan2015radiation}.
    
    \item \textit{Total column ozone} ( kg/m$^{2}$): is the total ozone along a column from the surface to the top of the atmosphere. It provides information on the densities in the atmosphere ~\cite{solomon2022stratospheric}.
\end{itemize}

 \subsection{Greenness index}
\label{subsec:dc_wildfires}

The last source of information used in this article is the greenness index (GI) or Green Leaf Index, which shows the relationship between the reflectance in the green channel compared to the other two visible light channels, red and blue ~\cite{gobron2000advanced}. Like the atmospheric variables they are geo-referenced and can be processed as images. The GI measurements were three times per month instead of daily, on the 1st, 11th and 21st day. The coordinate system used in this case is UTM (Universal Transverse Mercator) and the distance between two contiguous latitudes is 25 km compared to 111 km for the atmospheric variables. 
\begin{equation}
    \text{Greenness index} = \frac{2Green - Blue - Red}{2Green + Blue + Red}
\end{equation}

 \subsection{Data preparation}
\label{subsec:dc_wildfires}

Once we have defined the different data sources that we are going to use, we have to apply early fusion techniques that allow us to take advantage of all the available information. All these variables are georeferenced with X and Y coordinates for each value, a two-dimensional matrix where the columns correspond with the longitudes and the rows to the latitudes. However, they present differences: the coordinate systems and spatial/temporal resolution; therefore to apply early fusion of modalities, we need to preprocess the relevant information.
For instance, the greenness index uses the UTM (Universal Transverse Mercator) system, while the atmospheric variables use the decimal coordinate system. For homogeneity, the GI variable is converted to decimal. Another difference between the information sources is the resolution, the atmospheric variables have a distance between the latitude values of 111 km while the GI variable has a distance of 25 km. However, we need the resolution of all variables to be the same to be able to combine both information sources, so we decided to change the resolution of atmospheric variables, to match the number of columns and rows match between the two variables and early fusion techniques can be applied. The last difference between them is the temporal resolution. Atmospheric variables are recorded on a daily basis and GI is recorded three times per month (from the 1st to the 10th, 11th to the 21st and 21st to the end of the month). Therefore, we decided to divide the variables into two different groups as follows:

\begin{itemize}
    \item \textbf{Daily}: these variables are 10 metre U wind component and 10 metre V wind component. They are measured daily at 12:00 or 18:00, depending on the variable. If both measurements are available, we choose the first one, but if we do not have a measurement at 12:00, the one at 18:00 is taken. The value measured on the day of the fire was used for these variables because they are directly related to the occurrence of wildfires, as used in the Fire Weather Index \cite{van1987development}.
    
    \item \textbf{Trend}: this category includes the greenness index, the dewpoint temperature of 2 m, evaporation, net surface solar radiation, net surface thermal radiation, downward surface thermal radiation, and total ozone of columns. The greenness index was not modified. For atmospheric variables the time-series discrete difference was calculated between the dates of measurement of the greenness index. The reasons for using the trend of these variables are the following. 
    \begin{itemize}
        \item \textit{2 metre dewpoint temperature and evaporation}: are metrics of water stress, so the trend measure will allow us to know if the area is losing water vapour or increasing water stress, which would favour the spread of the fire.
        
        \item \textit{Surface net solar radiation and surface net thermal radiation} : measures the radiation emitted by the Earth's surface. If these variables increase, it means that the surface is getting warmer.
        
        \item \textit{Surface thermal radiation downwards}: is a measure of the thermal radiation that reaches the earth's surface. If this variable increases, it means that the daily temperature tends to increase, which increases evaporation and favours the spread of the fire.
        
        \item \textit{Total columns ozone}: this gas is in the atmosphere and absorbs solar radiation and prevents the temperature from rising. If the amount of ozone decreases over time, this means that more solar radiation reaches the surface and will tend to increase the temperature.
    \end{itemize}
    
\end{itemize}

To obtain all the matrices explained above, we take the wildfire central and extract the neighbouring data. Each set of fire coordinates results in a 128x128 matrix for each variable. We overlapped the different matrices generated at the end samples with dimensions 128x128x9. The labels are a list of six values, corresponding to the wildfire data records. Finally, the data were normalised, for the samples the z-score normalisation was used in all subsets (labelled and unlabelled samples) and for the labels the min-max normalisation was used.

\subsection{Data}
\label{subsec:dc_data}

As explained in section \ref{subsec:dc_wildfires}, the number of wildfire data records is limited, which means that the set of available labelled samples is limited, especially for Deep Learning, so we decided to create also two unlabelled subsets and two labelled corresponding to the two autonomous regions for which we have wildfire data records.
\begin{itemize}
    \item \textbf{Unlabelled subsets}: that corresponds with the subsets for which we do not have labels.
    
    \begin{itemize}
        \item \textit{Self-supervised data}: We have generated a set of 100,000 samples taken at random coordinates and times within the parameters of our study in Castilla y León to train the encoder model, as it would be impossible to train the model with the available labelled data.
        \item \textit{Map data}: this system predicts the different labels for a specific point, which is not sufficiently explanatory to apply it in real cases, so we have created a set of unlabelled samples covering the entire map of the autonomous region of Castilla y León to create a visualisation that shows the predictions for a specific day, which means that we will generate one visualisation per label. To create the prediction maps, in the case of Castilla y León, we have generated a total of 2,970,000 individual samples superimposed on each other to obtain predictions for each location on the map. 
    \end{itemize}
    
    \item \textbf{Labelled sets}: that corresponds to the subsets for which we have recorded the labels, as we have explained above, it corresponds to the measurements collected in both autonomous regions.
    \begin{itemize}
        \item \textit{Castilla y León}: set of 445 labelled samples from the autonomous region; we use this set to test the autoencoder, to fine-tune and test the WAM model representing, respectively, 70 and 30\% of the samples.
        \item \textit{Andalucía}: set of 151 labelled samples from Andalucía, we use it to test the WAM model. This set of samples allows us to test the generalisability of our system when we change the study area.
    \end{itemize}
\end{itemize}

\section{Methodology}
\label{section:methodology}

\begin{figure}[!h]
\centering
  \includegraphics[width=\textwidth, scale=1]{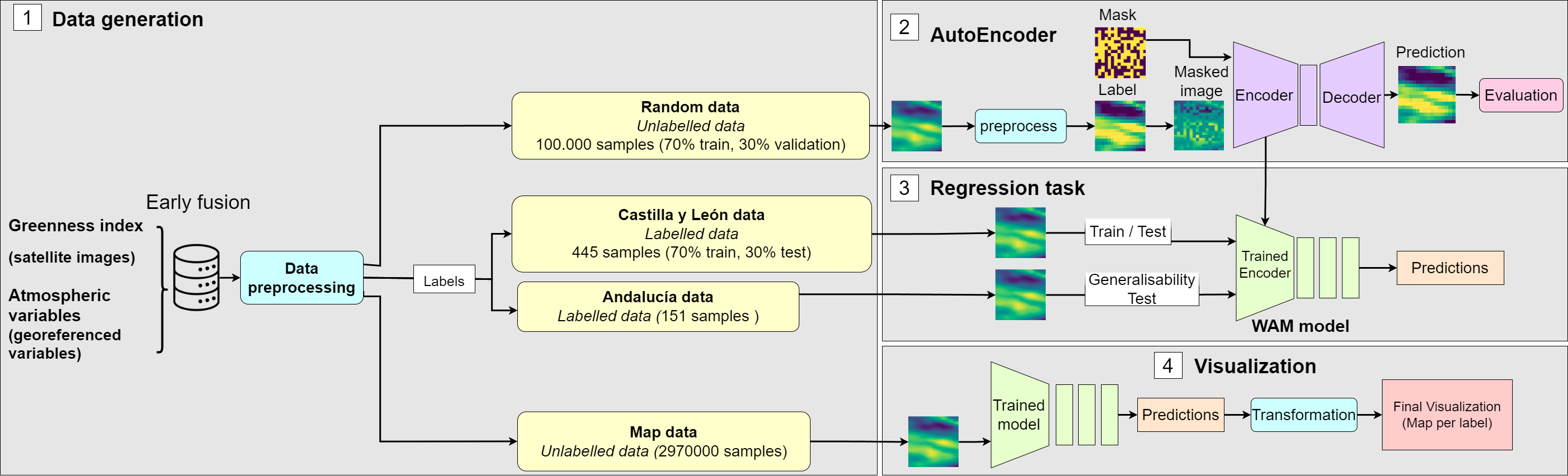}
  \caption{Visual representation of the proposed methodology for wildfire management. 1) \textit{Data generation}, from the different data sources, the different subsets (labelled and unlabelled) were generated; 2) \textit{Autoencoder}, training and validation of the autoencoder that learns the patterns of different atmospheric variables and greenness index; 3) \textit{Regression task}, development of the WAM model for the prediction of resources needed in fire management; 4) \textit{Visualization}, creation of daily prediction maps showing the resources needed if a fire were to break out at a particular point in the study area.}
  \label{fig:methodology}
\end{figure}

We can summarise the proposed methodology in Figure \ref{fig:methodology}, which is composed of four modules. The first one is the data preprocessing step, where we prepare the images and the labels for the different models. In the second one, we build the AutoEncoder, train it with the random dataset and test it with the dataset of Castilla y León. Then, we used the trained encoder to develop the regression model to estimate the wildfire cost variables. Using the regression model, we develop a map for each variable, where each pixel value corresponds to a prediction.

\subsection{Autoencoder pretraining}
\label{methodology:autoencoder_architecture}

\begin{figure}[!h]
\centering
  \includegraphics[width=3.00in, scale=1]{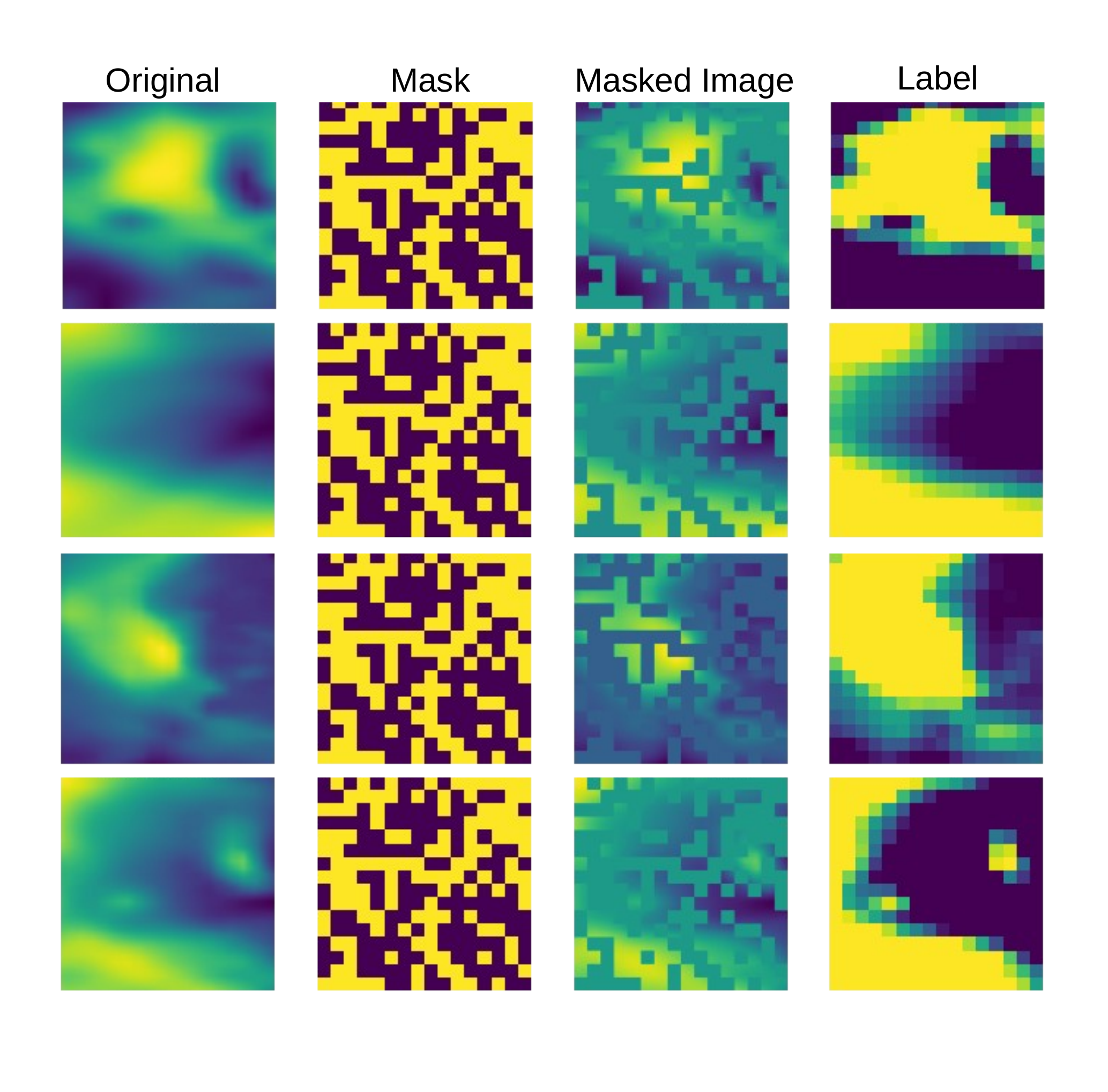}
  \caption{Example of the preprocessing showing four channels of a sample, where the first column represents the atmospheric variables; the second column represents the mask for that channel; the third represents masked image and finally the fourth sample represents the label for the autoencoder.}
  \label{fig:sample_preprocess}
\end{figure}

We developed an autoencoder for meteorological understanding. We approach this problem with a self-supervised masked image modeling (MIM) objective. We divide the 128x128 matrix into a grid of 8x8 patches where each patch is masked with a 0.5 probability (Figure \ref{fig:sample_preprocess} second column). Using the unmasked matrix patches (Figure \ref{fig:sample_preprocess} third column) the system tries to predict the average of the masked values for each masked patch (Figure \ref{fig:sample_preprocess} last column). In most papers such as ~\cite{he2022masked, feichtenhofer2022masked, pang2022masked, kim2022fast} the autoencoders attempt to reconstruct the original image directly. In our case, the autoencoder tries to learn to recognise the patterns and trends in the different channels (atmospheric variables), we do not consider a full reconstruction useful enough to warrant the added modelling complexity. For this reason, we assigned discrete categorical labels to the mean values of the patches. 

To select the adequate number of bins used in the experiments, we decided to run preliminary experiments with different values of bins: 4, 8, 16, 32 and 64 categories. We also explored for each experiment different values of learning rates (5e-5, 1e-4, 2e-4, 5e-4, 1e-3) in order to select the combination with the best performance on this task.
 

After preliminary experiments we settled on the following hyper-parameters for WAM pretraining as summarized in Table \ref{table:hyperparameters} 

\begin{table}[h]
\centering
\caption{Summary of parameters used in pretrain: optimization and training methodology.}
\label{table:hyperparameters}
\begin{tabular}{rl}
\hline
\textbf{Optimization}         &                                   \\ \hline
Optimizer                     & Adam                              \\
Learning rate                 & 1e-4                              \\
Loss function                 & Sparse Categorical Cross Entropy \\
Metric                        & Sparse Categorical Accuracy       \\
                              &                                   \\
\textbf{Training methodology} &                                   \\ \hline
Maximum epochs                & 2000                              \\
Checkpoint monitor            & Sparse Categorical Accuracy       \\
Batch size                    & 64                                \\
Image size                    & 128 x 128                         \\ 
\# bins                        & 64                                \\ 
Patch size                    & 16x16                             \\ \hline
\end{tabular}
\end{table}

\paragraph{Encoder architecture}

As we explained before, the encoder maps the features from the input into a latent representation. In this paper, we consider two different encoder architectures, with the purpose of developing an encoder that can better understand the patterns and trends of the variables. 
\begin{itemize}
    \item \textit{Sequential architecture}: is composed of three different convolutional blocks, see Figure \ref{fig:arch1}. Each has a convolutional layer with 128, 256, and 512, respectively, in the different convolutional blocks, with a kernel size of (3,3) and we apply the "same" padding method to adjust the size of the input to our requirement; the second is BatchNormalization layer followed by the activation function, which is ReLu; and finally there is a MaxPooling2D layer with a pool size of (2,2). Except for the convolutional layer, which has different values depending on the block, the rest have the same values in all convolutional blocks.

\begin{figure}[!h]
\centering
  \includegraphics[width=3.00in, scale=1]{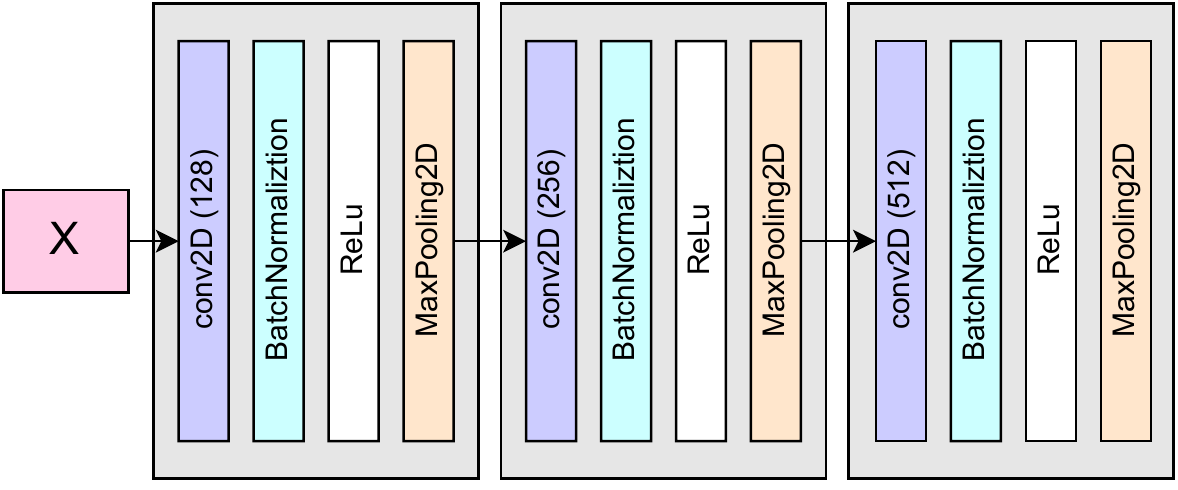}
  \caption{Visual representation of architecture 1, the different convolutional blocks can be seen in grey colour.}
  \label{fig:arch1}
\end{figure}

    \item \textit{Residual architecture}: In contrast to the previous architecture, this one presents skip connections or shortcuts to jump over some layers, see Figure \ref{fig:arch2}. These connections allow for deeper networks with less vanishing gradient issues. Our architecture is composed of three different convolutional blocks, as the previous one. Each block has four convolutional layers with the same number of neurons (128, 256 and 512, respectively) with a batch normalization layer and a Gelu activation layer; at the end of each convolutional block there is a Max pooling layer. The convolutional layer has a kernel size of (3,3) with padding. In each convolutional block, the activation layer receives the output of the previous layer and the output of the last activation one, but there are no connections between the different convolutional blocks.

\begin{figure}[!h]
\centering
  \includegraphics[width=\textwidth, scale=1]{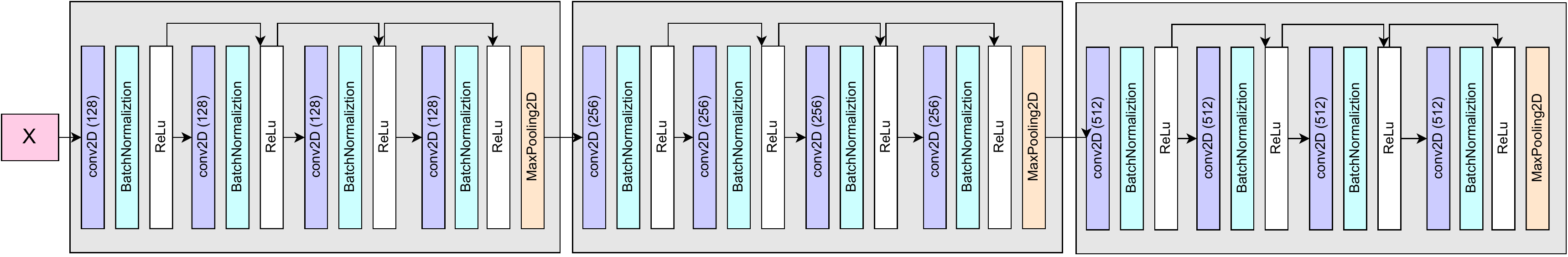}
  \caption{Visual representation of architecture 2, the different convolutional blocks can be seen in grey colour and the skip connections are represented by arrows.}
  \label{fig:arch2}
\end{figure}
\end{itemize}


\paragraph{Decoder architecture}

The decoder reconstructs the original image using the latent representation, i.e. filling in the masked patches. In other words, the decoder has to return the categorical values of the patches. Its architecture is composed of only two layers; the first one is a dense layer whose number of neurons is equal to the number of channels of the image, in our case 9, multiplied by the number of categories that the patches can take, 64, so the total number of neurons of the dense layer is 576, then there is a reshape layer which is in charge of changing the dimensions of the output; the desired dimension is (number of patches, number of patches, number of channels, categories). This allows the system to return the image with the same dimensions as the input. 

\subsection{The Wildfire Assessment (WAM) model}
\label{methodology:regression_model}
We assume that wildfires are inevitable and unpredictable but their spread is determined by natural causes and mitigated by human action. As such we build a model that can estimate the damage that a wildfire could provoke if it began on a given date; allowing for human experts to estimate required action. The labels to be used are the area that would burn in the fire, the time it would take to control and extinguish the wildfire, and the human, vehicle, and aerial resources needed to extinguish it. The number of samples is very small for deep learning techniques so we use a transfer learning approach, using the pre-trained weights from the previous task to fine-tune a regression model.

The latent representation is flattened and then perturbed by a Dropout with a rate of 70\%, which is passed to a dense layer with 512 neurons and GELU activation. Finally, the classification layer of the model consists of a dense layer with six neurons with linear activation. The model was trained for a maximum of 6,
000 epochs using the Adam optimiser with a learning rate of $1e-5$. The metric and the loss function used to control the evolution of the model were the mean absolute error (MAE) and the mean squared error (MSE), respectively, from which MAE was selected as a checkpoint to store the model weights. To train the regression model we used the dataset from Castilla y León, composed by 446 samples; and we used the dataset from Andalucía to test the generalization of the pre-training model to other areas with different environmental conditions.

\subsection{Baselines}

In this paper, as we have explained in Section \ref{methodology:visualization}, we have used a private dataset that has not been published yet, which makes it impossible to compare it with other papers that work with our dataset. Furthermore, we decided to treat the atmospheric variables as two-dimensional matrices, where the central point coincides with the geographical position of the fire, but also took into account the area around the point, which is a novel approach in the field. The articles found in the field use classical machine learning to solve their tasks; whether they predict susceptibility, the area that will burn in case of wildfires, among others. To offer a fair comparison with state of the art techniques, the selected techniques were extracted from reviews of the literature of work in this field~\cite{jain2020review, abid2021survey, bot2022systematic}: Decision trees, GBoost, Random forests, Support vector regression, and XGBoost.

Using the same arrays for the deep learning model we extract the mean, standard deviation and the centre point generating an array of 27 values for each sample. The matrices are generated from the original samples which, as mentioned above, had z-score normalization applied to them and were analysed using the techniques described above. On the other hand, as in the Deep Learning models, Min-Max normalization is applied to the labels, so to measure the real performance of the results obtained with the regression models, the predictions were first denormalized and the MAE was measured with the real labels and the denormalized predictions.

\subsection{Visualization}
\label{methodology:visualization}

The last step of the methodology of this manuscript is the visualization of a map of the autonomous region for each unit of time and label (burnt area, control time, extinction time, human resources, vehicle resources, and aerial resources). For each pixel of the map, a sample was generated with input dimensions explained in Section \ref{subsec:dc_wildfires}, a total of 2,970,000 samples are prepared to generate the map, a frame remains on the outside of the map that cannot be generated due to the input size requirement. The 6 predictions were extracted from each of the samples and structured in a two-dimensional matrix to reconstruct the maps, one per each label. 

\section{Results}
\label{section:results}

This section describes the results obtained with the proposed methodology and assesses its performance on the dataset explained in section \ref{section:data_collection}, for the autonomous regions of Castilla y León and Andalucía between 2001 and 2012. The results can be divided into four different sections: firstly we evaluate the autoencoder performance, where we first perform a test to choose the training parameters, and secondly we compare the results obtained for the two proposed architectures, using the Sparse Categorical Accuracy metric; secondly we evaluate the performance of the proposed regression model and compare it with the most used techniques for similar state-of-the-art problems, using the Mean Square Error (MAE) metric. The third section corresponds to the prediction map, and the last one corresponds to the known limitations.

\subsection{Performance of Autoencoder models}

As mentioned above, before training the final models we did an initial hyper-parameter search before pre-training. Our objective with this search is not to maximize accuracy, instead we search for the hardest objective the model can still solve with fair accuracy. As explained in section \ref{section:data_collection}, we combined different numbers of categories into which we divided the pixel values and different learning rate values. The results can be seen in Table \ref{table:explore_parameters_autoencoder}. As a limited number of epochs are used models with lower number of bins achieve better accuracy, however it is observed that all models are able to learn patterns and fill the hidden patches. Given the results obtained for the combination of learning rate = 1e-4 and 64 bins, which achieved an accuracy of 0.6, we chose these parameters to train the models. 

After the learning rate and number of classes are selected, the following experiments we will use a larger number of epochs and a larger batch size. This model was chosen because the high number of bins allows the model to learn fine-grained patterns and trends of the different atmospheric variables resulting in a better intermediate representation of the input data. In addition, the model trained with a learning rate of 1e-4 is the best performing model. 

\begin{table}[H]
\caption{Accuracy of each parameters combination in the AE training.}
\label{table:explore_parameters_autoencoder}
\centering
\begin{tabular}{rlllll}
\toprule
                                                        & \multicolumn{5}{c}{\textit{}{Number of bins}}     \\ 
\textit{Learning rate} & \textit{4}     & \textit{8}     & \textit{16}    & \textit{32}    & \textit{64}    \\ \midrule
1e-4                                                    & {0,975} & {0,939} & {0,852} & {0,736} & 0,600 \\
5e-5                                                    & {0,975} & {0,946} & {0,893} & {0,786} & 0,528 \\
2e-4                                                    & {0,966} & {0,949} & {0,876} & {0,731} & 0,567 \\
5e-4                                                    & {0,969} & {0,939} & {0,874} & {0,767} & 0,542 \\
1e-3                                                    & {0,968} & {0,949} & {0,878} & {0,759} & 0,558 \\ \bottomrule
\end{tabular}
\end{table}

\begin{figure}[!h]
\centering
  \includegraphics[width=0.5\textwidth]{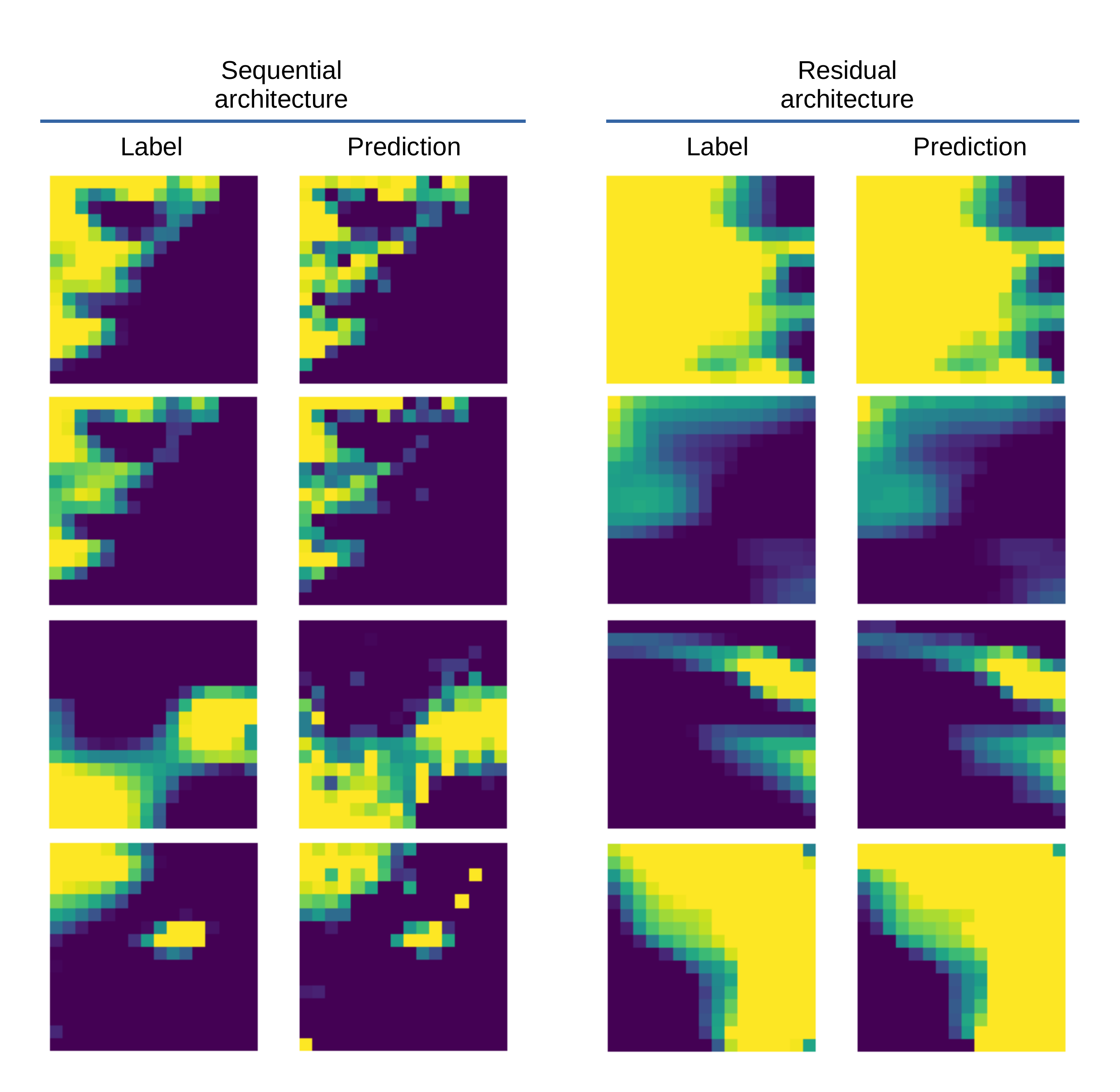}
  \caption{Examples of results from the two AutoEncoder for different samples of the labelled training set. The first and third columns correspond to the AutoEncoder labels, the second and fourth columns are the predictions of the two proposed architectures, sequential and residual architectures.}
  \label{fig:examples_autoencoder}
\end{figure}

Once we select the training parameters, we train the proposed architectures: sequential and residual ones. Table \ref{table:autoencoder_results} shows the results obtained for the labelled training dataset, as it was trained and validated with the unlabelled random dataset. We can observe how the residual architecture results obtained better results, 0.861, compared to the sequential one, which achieved results of 0.773. Figure \ref{fig:examples_autoencoder} confirms these results, as we can see that the residual one successfully reconstructs an image from the latent representation. Although there are differences between the results of both architectures, we decided to create regression models with both encoders to test their performance.

\begin{table}[!h]
\centering
\caption{Results of the two AutoEncoder architectures evaluated with the labelled training set.}
\label{table:autoencoder_results}
\begin{tabular}{rl}
\toprule
& \textit{Accuracy} \\ 
Sequential architecture & 0.773    \\
Residual architecture & \textbf{0.861}    \\ \bottomrule
\end{tabular}
\end{table}

The results obtained, Figure \ref{fig:examples_autoencoder} and Table \ref{table:autoencoder_results}, show that the autoencoder models are able to understand the patterns and trends of different variables of the atmospheric variables and greenness index. It can be seen that the residual architecture obtains better results than the sequential architecture. The differences between the two architectures are the number of convolutional layers in each block and the use of skip connection or shortcuts. The residual architecture has more trainable parameters and the skip connections help avoid vanishing gradient issues, therefore this behaviour is within expectations.

\subsection{WAM model}
We analyze the performance of the different regression models proposed, table \ref{table:regression_results}. We use the two encoders generated in the previous step and tested with and without fine-tuning. To check if these models work correctly, we compare them with five different baselines. Other deep learning techniques are not available at the time of writing, thus we choose the most common state-of-the-art techniques to compare with. Firstly, we observe that the worst performing model is Support Vector Regression, which presents the highest MAE values in all classes except aerial resources and control time; the second worst baseline is the decision tree, which presents the worst results in control time and aerial resources. The best performing baselines are GBoost and random forest, which improve the results of the models with frozen encoder for the human and heavy resource classes. All baselines predictions are better than the average of the training values. 

We also compare our techniques to the baselines. We observe that the sequential architecture sometimes fails to achieve better performance than the baselines while in the other three classes (burnt surface, control time, and extinguishing time) they behave similarly to the fine-tuned encoder. On the other hand we observe that fine-tuning the entire encoder to the task achieves on average the best performance. Finally, if we compare the results obtained by the regression models generated with the encoder from sequential and residual architectures with fine-tune encoder, we can see that residual architecture works better, presenting the lowest MAE for three classes, burnt area, control time, and aerial resources; that is, it is the model with the best results of the nine models presented in table \ref{table:regression_results}.

The results obtained in this section show how deep learning techniques can outperform classical machine learning models, which are used in the state of the art in the field of fire management. Furthermore, we can see how few-shot learning techniques achieve very good results in fire management with a limited number of samples. Our fine-tuned residual encoder has shown to achieve better to the rest of the models and baselines shown, although there is still a great margin for improvement, the results are promising.

\begin{table}[!h]
\caption{Results of the four regression models generated, frozen encoder and fine-tune encoder from both architectures, compared with the average and five models from the state of the art. Best result in bold.}
\label{table:regression_results}
\resizebox{\textwidth}{!}{
\begin{tabular}{rllllll|llll|l}
\toprule
\multicolumn{1}{l}{}                                                & \multicolumn{1}{c}{\multirow{2}{*}{\begin{tabular}[c]{@{}c@{}}Average\\ baseline\end{tabular}}} & \multicolumn{1}{c}{\multirow{2}{*}{\begin{tabular}[c]{@{}c@{}}Decission\\ Tree\end{tabular}}} & \multicolumn{1}{c}{\multirow{2}{*}{GBoost}} & \multicolumn{1}{c}{\multirow{2}{*}{\begin{tabular}[c]{@{}c@{}}Random \\ Forest\end{tabular}}} & \multicolumn{1}{c}{\multirow{2}{*}{\begin{tabular}[c]{@{}c@{}}Support\\ Vector\\ Regression\end{tabular}}} & \multicolumn{1}{c}{\multirow{2}{*}{XGBoost}} & \multicolumn{2}{|c}{\begin{tabular}[c]{@{}c@{}}Frozen\\ encoder\end{tabular}}                                                                                                 & \multicolumn{2}{c|}{\begin{tabular}[c]{@{}c@{}}Fine-tune\\ encoder\end{tabular}}                                                                                             & \multicolumn{1}{c}{\multirow{2}{*}{\begin{tabular}[c]{@{}c@{}}Improvement\\ (\%)\end{tabular}}} \\ \cline{8-11}
\multicolumn{1}{l}{}                                                & \multicolumn{1}{c}{}                                                                            & \multicolumn{1}{c}{}                                                                          & \multicolumn{1}{c}{}                        & \multicolumn{1}{c}{}                                                                          & \multicolumn{1}{c}{}                                                                                       & \multicolumn{1}{c}{}                         & \multicolumn{1}{|c}{\begin{tabular}[c]{@{}c@{}}Sequential \\ architecture\end{tabular}} & \multicolumn{1}{c}{\begin{tabular}[c]{@{}c@{}}Residual\\ architecture\end{tabular}} & \multicolumn{1}{c}{\begin{tabular}[c]{@{}c@{}}Sequential\\ architecture\end{tabular}} & \multicolumn{1}{c|}{\begin{tabular}[c]{@{}c@{}}Residual\\ architecture\end{tabular}} & \multicolumn{1}{c}{}                                                                            \\ \midrule
\begin{tabular}[c]{@{}r@{}}Burnt \\ Area (m)\end{tabular}           & 406,189                                                                                         & 434,473                                                                                       & 333,814                                     & 287,051                                                                              & 747,340                                                                                                    & 302,683                                      & 280,800                                                                                & 255,800                                                                             & 278,200                                                                               & \textbf{233,100}                                                                             & 18,8\%                                                                                          \\
\begin{tabular}[c]{@{}r@{}}Control Time \\ (min)\end{tabular}       & 1846,537                                                                                        & 1875,236                                                                                      & 1274,760                           & 1416,345                                                                                      & 1660,959                                                                                                   & 1295,846                                     & 1218,000                                                                               & 1212,000                                                                            & 1192,000                                                                              & \textbf{1114,000}                                                                            & 21\%                                                                                            \\
\begin{tabular}[c]{@{}r@{}}Extintion Time \\ (min)\end{tabular}     & 3255,813                                                                                        & 2566,528                                                                                      & 2630,425                                    & 2538,686                                                                             & 2797,823                                                                                                   & 2654,167                                     & 2234,000                                                                               & 2338,000                                                                            & \textbf{2230,000}                                                                              & 2280,000                                                                            & 10,2\%                                                                                          \\
\begin{tabular}[c]{@{}r@{}}Human \\ resources (units)\end{tabular}  & 87,582                                                                                          & 66,289                                                                                        & 54,785                                      & 53,373                                                                               & 82,152                                                                                                     & 57,140                                       & 73,700                                                                                 & \textbf{51,100}                                                                              & 73,100                                                                                & 52,600                                                                              & 1,4\%                                                                                           \\
\begin{tabular}[c]{@{}r@{}}Heavy \\ resources (units)\end{tabular}  & 6,047                                                                                           & 5,056                                                                                         & 4,139                                       & 4,011                                                                                & 6,450                                                                                                      & 4,528                                        & 5,918                                                                                  & \textbf{3,672}                                                                               & 5,930                                                                                 & 3,863                                                                               & 3,7\%                                                                                           \\
\begin{tabular}[c]{@{}r@{}}Aerial \\ resources (units)\end{tabular} & 5,197                                                                                           & 4,309                                                                                         & 3,283                                       & 3,169                                                                                & 3,626                                                                                                      & 3,318                                        & 3,684                                                                                  & 2,988                                                                               & 3,744                                                                                 & \textbf{2,883}                                                                               & 9\%                  \\ \bottomrule                                                                          
\end{tabular}}
\end{table}

\subsection{Visualization}

\begin{figure}[!h]%
\centering%
  \includegraphics[width=3.00in, scale=1]{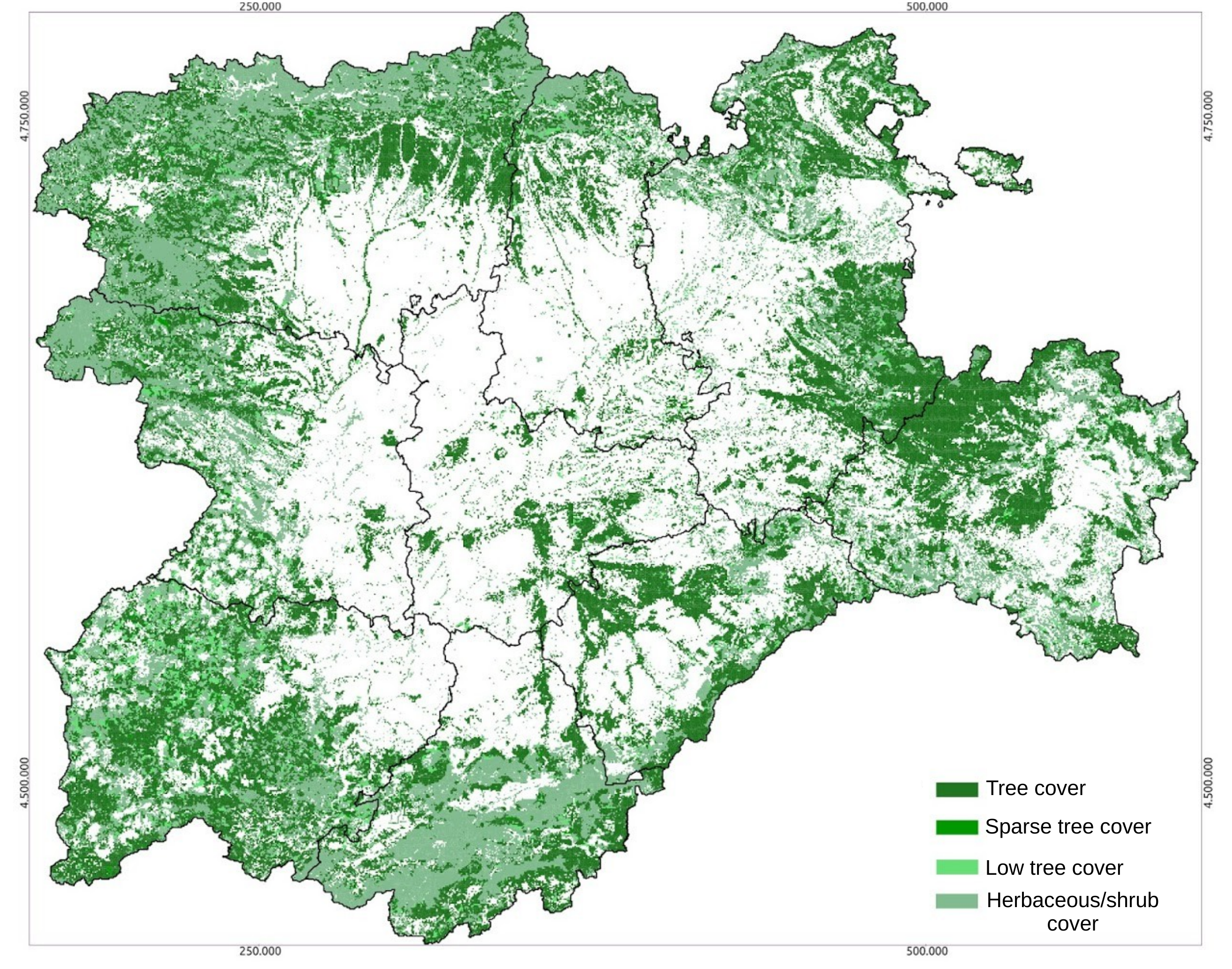}%
  \caption[]{%
  Forest map of Castilla y León. Dark green: Tree cover; Green: Sparse tree cover; light green: low tree cover; and grey\-green: herbaceous\-shrub cover. This map is an adaptation of the forest map provided by Ministerio para la Transición Ecológica y el Reto Demográfico (Spain)\footnotemark%
  }%

  \label{fig:forest_map}%
\end{figure}%

\footnotetext{\url{https://www.miteco.gob.es/es/}}

\begin{figure*}[!h]
        \subfloat[Burnt Area]{%
            \includegraphics[width=.32\linewidth]{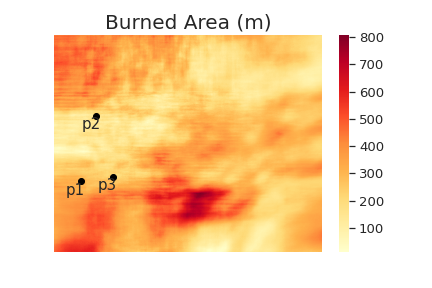}%
            \label{subfig:a}%
        }\hfill
        \subfloat[Control time]{%
            \includegraphics[width=.32\linewidth]{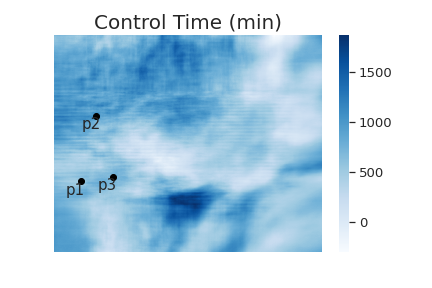}%
            \label{subfig:b}%
        } \hfill
        \subfloat[Extinction time]{%
            \includegraphics[width=.32\linewidth]{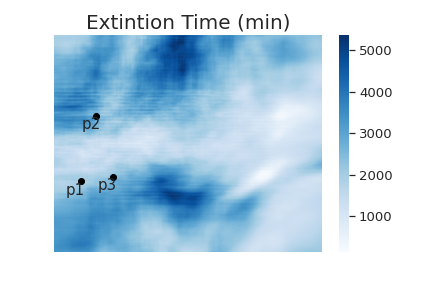}%
            \label{subfig:c}%
        }\\
        \subfloat[Personal resources]{%
            \includegraphics[width=.32\linewidth]{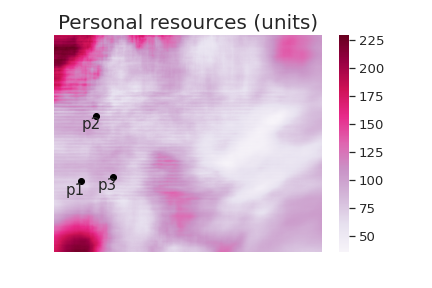}%
            \label{subfig:d}%
        }\hfill
        \subfloat[Heavy resources]{%
            \includegraphics[width=.32\linewidth]{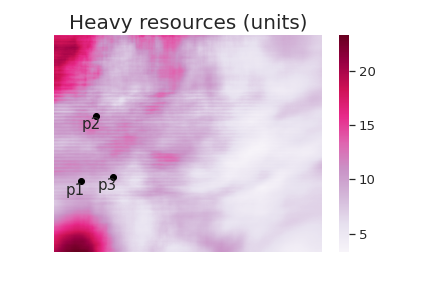}%
            \label{subfig:c}%
        }\hfill
        \subfloat[Aerial resources]{%
            \includegraphics[width=.32\linewidth]{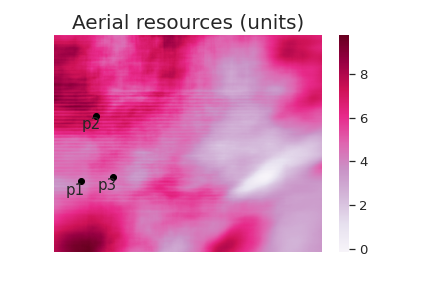}%
            \label{subfig:d}%
        }
        \caption{Prediction maps, a visualization of burnt area, control and extinction time and needed resources. }
        \label{fig:prediction_maps}
    \end{figure*}

As we explain in Section \ref{section:methodology}, in addition to testing our model in the test set, to analyse its operation and performance, we have created a prediction map for a specific date, with three test samples. With these visualizations we want to analyse the risk assessment throughout the autonomous region when a fire occurs. As shown in figure \ref{fig:prediction_maps}, we have created a map for each of the labels, showing the resources that would be needed in case of fire.

In all maps we can see that the areas close to the black spots, which correspond to the wildfires that occurred on a specific date, are the ones that would need more resources. If we compare these results with forest and wildfire maps, Figures \ref{fig:map_cyl_and} and \ref{fig:forest_map}, respectively, firstly, we can see that there are three wildfires in the west of the autonomous region, and also the areas with the highest values for the six labels are close to these points. In addition, these areas coincide in many areas with the highest vegetation cover. The evaluation of such visualizations is not straightforward, as we do not have a ground truth to compare with, but we can check if the marked areas match with the expected ones, i.e. if they are close to the fires.

We do not want to know exactly where a fire is, but we can give indications of how damaging a fire could be to an area. High hazard areas are high hazard areas because they have conditions that are prone to fire spread, which is an indicator that there is a greater likelihood of a fire occurring in that area.

\subsection{Known limitations}

As we explained before in Section \ref{section:data_collection}, we used the Andalucía dataset to check the generalization of the model without fine-tuning with samples from the new area with different environmental conditions. For this purpose, we use the regression model that obtained the best performance in the Castilla y León dataset, the fine-tuned residual architecture encoder. Like in the previous section, we compare our model with five different baselines and the average. 

First of all, we can observe that the baselines results are similar in both experiments, except for the last label, aerial resources, where all baselines perform worse than the average baseline. Second, it is notable that the support vector regression and XGBoost models obtain the same results for both datasets. The rest of the models for the other five labels achieve better results than the average one. About our model only two labels have a lower MAE than average one: control and extinction time. The other variables have a higher MAE than average ones, burnt area, human, heavy and aerial resources. It is important to note that there are some other factors, such as: distance between the water sources and the fire location, type of plane or the orography, that may have an influence on the resource prediction. 

The model we have used to test the new samples from Andalucía was fine-tuned with a set of 446 samples from Castilla y León, a limited region with different meteorological features and vegetation than Andalucía, so the model is not able to correctly predict the resources that would be needed in case of fires, except for the control and extinction time classes. The model has been trained by means of the atmospheric variables in a region surrounding Castilla y León region. Thus, different regions are affected by changes in the meteorological conditions. The model may need to be adjusted if it is to be applied to other areas, i.e. pretraining with a small sample set of the area of interest. Another factor that could also affect its generalization is that the encoder was trained with a set of random samples from Castilla y León.

\begin{table}[H]
\caption{Results of residual architecture with fine-tune encoder and baselines for Andalucía dataset.}
\label{table:regression_results}
\centering
\resizebox{0.8\textwidth}{!}{
\begin{tabular}{rlllllll}
\toprule
                         & \begin{tabular}[c]{@{}c@{}} \\Average\\baseline\end{tabular}     & \begin{tabular}[c]{@{}c@{}} \\Decision\\ Tree\end{tabular} & \begin{tabular}[c]{@{}c@{}} \\ \\ GBoost\end{tabular}   & \begin{tabular}{@{}c@{}} \\ Random \\ Forest \end{tabular} & \begin{tabular}{@{}c@{}}Support \\ Vector \\ Regression\end{tabular} & \begin{tabular}[c]{@{}c@{}} \\ \\ XGBoost\end{tabular}  & \begin{tabular}[c]{@{}c@{}} \\ \\ WAM\end{tabular} \\ \midrule
Burnt area (m)          & 393.209  & 391.556      & 320.148  & \textbf{293.271}      & 747.340                 & 302.683  & 751.5     \\
Control Time (min)       & 1747.570 & 1713.208     & 1276.427 & 1371.579     & 1660.959                & 1295.846 & \textbf{950.0}     \\
Extintion Time (min)     & 3067.964 & 2798.904     & 2559.446 & 2523.677     & 2797.823                & 2654.167 & \textbf{2268.0}    \\
Human resources (units)  & 83.971   & 63.671       & 56.237   & \textbf{53.931}       & 82.152                  & 57.140   & 206.6     \\
Heavy resources (units)  & 5.996    & 5.247        & 4.189    & \textbf{4.061}        & 6.450                   & 4.528    & 14.414    \\
Aerial resources (units) & 4.993    & 6.107        & 5.817    & \textbf{5.665}        & 6.103                   & 5.918    & 11.01     \\ \bottomrule
\end{tabular}}
\end{table}

\section{Conclusions and future work}
\label{section:conclusion_future_work}

This paper proposes the WAM model architecture for the prediction of wildfire resource using atmospheric variables and the greenness index. Due to the limited number of labelled samples, we trained an autoencoder with unlabelled samples to learn the trends or patterns of the variables of a geographical area and apply the knowledge to a regression task that predicts the resources needed, the control and extinguishing time, and the area that would burn in case of a fire. Finally, we tested its ability to generalise to other areas with different meteorological conditions, using a dataset from Andalucía. The application in a specific location is limited, for this reason we create a prediction map for each label, showing the necessary resources in each location of the autonomous region of Castilla y León for a specific date. 

The results are promising and the methodology is novel in the field of wildfires, as most published papers use machine learning algorithms without exploring other Deep Learning options or techniques. Firstly, the residual autoencoder architecture presents higher performance than the sequential one, because the skip connection provides an alternative path for the gradient with backpropagation and allows deeper layers to learn from the information of initial layers. However, it can be observed that better results are achieved by using the residual autoencoder architecture if the encoder is retrained in the regression task. Possibly the addition of new samples allows the regression model to pick up patterns that the autoencoder was not able to learn originally.

The model we propose is trained in a very specific region, an autonomous region of Spain, Castilla y León, so we analyse its generalization to other areas with different meteorological conditions, in our case Andalucía, another autonomous region of the same country. The results show that our model is able to correctly predict two of the labels, control and extinguishing time, but not the rest of the labels, i.e. WAM model is not able to adapt to other regions without retraining. As we have explained above, the WAM model obtained better results when the encoder was retrained in the regression task, although the samples belonged to the same autonomous region. As the original model was pre-trained on a specific region it is not able to generalise to other areas with different meteorological conditions. 

We have identified clear weaknesses of the model that can be sorted in several ways. First, the generalization capabilities fall short due to the low amount and specialization of pretraining data. This can be solved by widening the scope of the semi-supervised dataset (for example, cover the entire peninsula) and sampling more data points from the selected region. To support the increase in the dataset, a proportional enhancement of the model would be required; while maintaining the training methodology, a more powerful architecture could be applied, such as transformers or ConvNeXT. Another way to improve the performance of the proposed system is including more variables, some are included in many state-of-the-art works, such as distance to the nearest population, substrate use or orography and other variables closely related to wildfires management such as fuel models. Finally, as we can see from the published works in the area of wildfire management, this is an underexplored domain with numerous possibilities.


\section*{Acknowledgement}

This work has been supported by the research project DisTrack: Tracking disinformation in Online Social Networks through Deep Natural Language Processing, granted by Mobile World Capital Foundation; by the Spanish Ministry of Science and Innovation under FightDIS (PID2020-117263GB-100); by MCIN/AEI/10.13039/501100011033/ and European Union NextGenerationEU/PRTR for XAI-Disinfodemics (PLEC2021-007681) grant, by Comunidad Autónoma de Madrid under S2018/TCS-4566 grant, by European Comission under IBERIFIER - Iberian Digital Media Research and Fact-Checking Hub (2020-EU-IA-0252); and by "Convenio Plurianual with the Universidad Politécnica de Madrid in the actuation line of Programa de Excelencia para el Profesorado Universitario"





\bibliographystyle{abbrvnat}
\bibliography{bibtex.bib}

\begin{thebibliography}{66}
\providecommand{\natexlab}[1]{#1}
\providecommand{\url}[1]{\texttt{#1}}
\expandafter\ifx\csname urlstyle\endcsname\relax
  \providecommand{\doi}[1]{doi: #1}\else
  \providecommand{\doi}{doi: \begingroup \urlstyle{rm}\Url}\fi

\bibitem[Abid(2021)]{abid2021survey}
F.~Abid.
\newblock A survey of machine learning algorithms based forest fires prediction
  and detection systems.
\newblock \emph{Fire Technology}, 57\penalty0 (2):\penalty0 559--590, 2021.

\bibitem[Al-Fugara et~al.(2021)Al-Fugara, Mabdeh, Ahmadlou, Pourghasemi,
  Al-Adamat, Pradhan, and Al-Shabeeb]{al2021wildland}
A.~Al-Fugara, A.~N. Mabdeh, M.~Ahmadlou, H.~R. Pourghasemi, R.~Al-Adamat,
  B.~Pradhan, and A.~R. Al-Shabeeb.
\newblock Wildland fire susceptibility mapping using support vector regression
  and adaptive neuro-fuzzy inference system-based whale optimization algorithm
  and simulated annealing.
\newblock \emph{ISPRS International Journal of Geo-Information}, 10\penalty0
  (6):\penalty0 382, 2021.

\bibitem[Alados et~al.(2003)Alados, Foyo-Moreno, Olmo, and
  Alados-Arboledas]{alados2003relationship}
I.~Alados, I.~Foyo-Moreno, F.~Olmo, and L.~Alados-Arboledas.
\newblock Relationship between net radiation and solar radiation for semi-arid
  shrub-land.
\newblock \emph{Agricultural and Forest Meteorology}, 116\penalty0
  (3-4):\penalty0 221--227, 2003.

\bibitem[Baars et~al.(2021)Baars, Radenz, Floutsi, Engelmann, Althausen, Heese,
  Ansmann, Flament, Dabas, Trapon, et~al.]{baars2021californian}
H.~Baars, M.~Radenz, A.~A. Floutsi, R.~Engelmann, D.~Althausen, B.~Heese,
  A.~Ansmann, T.~Flament, A.~Dabas, D.~Trapon, et~al.
\newblock Californian wildfire smoke over europe: A first example of the
  aerosol observing capabilities of aeolus compared to ground-based lidar.
\newblock \emph{Geophysical Research Letters}, 48\penalty0 (8):\penalty0
  e2020GL092194, 2021.

\bibitem[Bonannella et~al.(2022)Bonannella, Chirici, Travaglini, Pecchi, Vangi,
  D’Amico, and Giannetti]{bonannella2022characterization}
C.~Bonannella, G.~Chirici, D.~Travaglini, M.~Pecchi, E.~Vangi, G.~D’Amico,
  and F.~Giannetti.
\newblock Characterization of wildfires and harvesting forest disturbances and
  recovery using landsat time series: A case study in mediterranean forests in
  central italy.
\newblock \emph{Fire}, 5\penalty0 (3):\penalty0 68, 2022.

\bibitem[Bot and Borges(2022)]{bot2022systematic}
K.~Bot and J.~G. Borges.
\newblock A systematic review of applications of machine learning techniques
  for wildfire management decision support.
\newblock \emph{Inventions}, 7\penalty0 (1):\penalty0 15, 2022.

\bibitem[Boulanger et~al.(2018)Boulanger, Parisien, and
  Wang]{boulanger2018model}
Y.~Boulanger, M.-A. Parisien, and X.~Wang.
\newblock Model-specification uncertainty in future area burned by wildfires in
  canada.
\newblock \emph{International journal of wildland fire}, 27\penalty0
  (3):\penalty0 164--175, 2018.

\bibitem[Buckland et~al.(2019)Buckland, Bailey, and Thomas]{buckland2019using}
C.~Buckland, R.~Bailey, and D.~Thomas.
\newblock Using artificial neural networks to predict future dryland responses
  to human and climate disturbances.
\newblock \emph{Scientific reports}, 9\penalty0 (1):\penalty0 1--13, 2019.

\bibitem[Campos-Vargas and Vargas-Sanabria(2021)]{campos2021assessing}
C.~Campos-Vargas and D.~Vargas-Sanabria.
\newblock Assessing the probability of wildfire occurrences in a neotropical
  dry forest.
\newblock \emph{{\'E}coscience}, 28\penalty0 (2):\penalty0 159--169, 2021.

\bibitem[Cansler et~al.(2022)Cansler, Kane, Hessburg, Kane, Jeronimo, Lutz,
  Povak, Churchill, and Larson]{cansler2022previous}
C.~A. Cansler, V.~R. Kane, P.~F. Hessburg, J.~T. Kane, S.~M. Jeronimo, J.~A.
  Lutz, N.~A. Povak, D.~J. Churchill, and A.~J. Larson.
\newblock Previous wildfires and management treatments moderate subsequent fire
  severity.
\newblock \emph{Forest Ecology and Management}, 504:\penalty0 119764, 2022.

\bibitem[Choi and Jun(2020)]{choi2020fire}
M.-Y. Choi and S.~Jun.
\newblock Fire risk assessment models using statistical machine learning and
  optimized risk indexing.
\newblock \emph{Applied Sciences}, 10\penalty0 (12):\penalty0 4199, 2020.

\bibitem[Coffield et~al.(2019)Coffield, Graff, Chen, Smyth, Foufoula-Georgiou,
  and Randerson]{coffield2019machine}
S.~R. Coffield, C.~A. Graff, Y.~Chen, P.~Smyth, E.~Foufoula-Georgiou, and J.~T.
  Randerson.
\newblock Machine learning to predict final fire size at the time of ignition.
\newblock \emph{International journal of wildland fire}, 28\penalty0
  (11):\penalty0 861--873, 2019.

\bibitem[Cornejo-Bueno et~al.()Cornejo-Bueno, P{\'e}rez-Aracil, Casanova-Mateo,
  Sanz-Justo, and Salcedo-Sanz]{cornejo4231494machine}
L.~M. Cornejo-Bueno, J.~P{\'e}rez-Aracil, C.~Casanova-Mateo, J.~Sanz-Justo, and
  S.~Salcedo-Sanz.
\newblock Machine learning classification-regression schemes for desert locust
  presence prediction in western africa.
\newblock \emph{Available at SSRN 4231494}.

\bibitem[Dhankar et~al.(2021)Dhankar, Gupta, et~al.]{dhankar2021systematic}
A.~Dhankar, N.~Gupta, et~al.
\newblock A systematic review of techniques, tools and applications of machine
  learning.
\newblock In \emph{2021 Third International Conference on Intelligent
  Communication Technologies and Virtual Mobile Networks (ICICV)}, pages
  764--768. IEEE, 2021.

\bibitem[Farasin et~al.(2020)Farasin, Colomba, and Garza]{farasin2020double}
A.~Farasin, L.~Colomba, and P.~Garza.
\newblock Double-step u-net: A deep learning-based approach for the estimation
  of wildfire damage severity through sentinel-2 satellite data.
\newblock \emph{Applied Sciences}, 10\penalty0 (12):\penalty0 4332, 2020.

\bibitem[Feichtenhofer et~al.(2022)Feichtenhofer, Fan, Li, and
  He]{feichtenhofer2022masked}
C.~Feichtenhofer, H.~Fan, Y.~Li, and K.~He.
\newblock Masked autoencoders as spatiotemporal learners.
\newblock \emph{arXiv preprint arXiv:2205.09113}, 2022.

\bibitem[Fern{\'a}ndez et~al.(2016)Fern{\'a}ndez, Fern{\'a}ndez,
  F{\'e}m{\'e}nias, and Peter]{fernandez2016copernicus}
J.~Fern{\'a}ndez, C.~Fern{\'a}ndez, P.~F{\'e}m{\'e}nias, and H.~Peter.
\newblock The copernicus sentinel-3 mission.
\newblock In \emph{ILRS workshop}, pages 1--4, 2016.

\bibitem[Fister et~al.(2022)Fister, P{\'e}rez-Aracil,
  Pel{\'a}ez-Rodr{\'\i}guez, Del~Ser, and Salcedo-Sanz]{fister2022accurate}
D.~Fister, J.~P{\'e}rez-Aracil, C.~Pel{\'a}ez-Rodr{\'\i}guez, J.~Del~Ser, and
  S.~Salcedo-Sanz.
\newblock Accurate long-term air temperature prediction with a fusion of
  artificial intelligence and data reduction techniques.
\newblock \emph{arXiv preprint arXiv:2209.15424}, 2022.

\bibitem[Gobron et~al.(2000)Gobron, Pinty, Verstraete, and
  Widlowski]{gobron2000advanced}
N.~Gobron, B.~Pinty, M.~M. Verstraete, and J.-L. Widlowski.
\newblock Advanced vegetation indices optimized for up-coming sensors: Design,
  performance, and applications.
\newblock \emph{IEEE Transactions on Geoscience and Remote Sensing},
  38\penalty0 (6):\penalty0 2489--2505, 2000.

\bibitem[Guz et~al.(2021)Guz, Gill, and Kulakowski]{guz2021long}
J.~Guz, N.~S. Gill, and D.~Kulakowski.
\newblock Long-term empirical evidence shows post-disturbance climate controls
  post-fire regeneration.
\newblock \emph{Journal of Ecology}, 109\penalty0 (12):\penalty0 4007--4024,
  2021.

\bibitem[Haiden et~al.(2018)Haiden, Sandu, Balsamo, Arduini, and
  Beljaars]{haiden2018addressing}
T.~Haiden, I.~Sandu, G.~Balsamo, G.~Arduini, and A.~Beljaars.
\newblock Addressing biases in near-surface forecasts.
\newblock \emph{ECMWF Newsletter}, 157:\penalty0 20--25, 2018.

\bibitem[He et~al.(2022)He, Chen, Xie, Li, Doll{\'a}r, and
  Girshick]{he2022masked}
K.~He, X.~Chen, S.~Xie, Y.~Li, P.~Doll{\'a}r, and R.~Girshick.
\newblock Masked autoencoders are scalable vision learners.
\newblock In \emph{Proceedings of the IEEE/CVF Conference on Computer Vision
  and Pattern Recognition}, pages 16000--16009, 2022.

\bibitem[Hersbach et~al.(2020)Hersbach, Bell, Berrisford, Hirahara,
  Hor{\'a}nyi, Mu{\~n}oz-Sabater, Nicolas, Peubey, Radu, Schepers,
  et~al.]{hersbach2020era5}
H.~Hersbach, B.~Bell, P.~Berrisford, S.~Hirahara, A.~Hor{\'a}nyi,
  J.~Mu{\~n}oz-Sabater, J.~Nicolas, C.~Peubey, R.~Radu, D.~Schepers, et~al.
\newblock The era5 global reanalysis.
\newblock \emph{Quarterly Journal of the Royal Meteorological Society},
  146\penalty0 (730):\penalty0 1999--2049, 2020.

\bibitem[Hodges and Lattimer(2019)]{hodges2019wildland}
J.~L. Hodges and B.~Y. Lattimer.
\newblock Wildland fire spread modeling using convolutional neural networks.
\newblock \emph{Fire technology}, 55\penalty0 (6):\penalty0 2115--2142, 2019.

\bibitem[Hogan(2015)]{hogan2015radiation}
R.~Hogan.
\newblock Radiation quantities in the ecmwf model and mars.
\newblock \emph{ECMWF, 2016}, 2015.

\bibitem[Jain et~al.(2020)Jain, Coogan, Subramanian, Crowley, Taylor, and
  Flannigan]{jain2020review}
P.~Jain, S.~C. Coogan, S.~G. Subramanian, M.~Crowley, S.~Taylor, and M.~D.
  Flannigan.
\newblock A review of machine learning applications in wildfire science and
  management.
\newblock \emph{Environmental Reviews}, 28\penalty0 (4):\penalty0 478--505,
  2020.

\bibitem[Janiec and Gadal(2020)]{janiec2020comparison}
P.~Janiec and S.~Gadal.
\newblock A comparison of two machine learning classification methods for
  remote sensing predictive modeling of the forest fire in the north-eastern
  siberia.
\newblock \emph{Remote Sensing}, 12\penalty0 (24):\penalty0 4157, 2020.

\bibitem[Jones et~al.(2022)Jones, Abatzoglou, Veraverbeke, Andela, Lasslop,
  Forkel, Smith, Burton, Betts, van~der Werf, et~al.]{jones2022global}
M.~W. Jones, J.~T. Abatzoglou, S.~Veraverbeke, N.~Andela, G.~Lasslop,
  M.~Forkel, A.~J. Smith, C.~Burton, R.~A. Betts, G.~R. van~der Werf, et~al.
\newblock Global and regional trends and drivers of fire under climate change.
\newblock \emph{Reviews of Geophysics}, 60\penalty0 (3):\penalty0
  e2020RG000726, 2022.

\bibitem[Julian and Kochenderfer(2018)]{julian2018autonomous}
K.~D. Julian and M.~J. Kochenderfer.
\newblock Autonomous distributed wildfire surveillance using deep reinforcement
  learning.
\newblock In \emph{2018 AIAA Guidance, Navigation, and Control Conference},
  page 1589, 2018.

\bibitem[Kang et~al.(2020)Kang, Jang, Im, Kwon, and Kim]{kang2020developing}
Y.~Kang, E.~Jang, J.~Im, C.~Kwon, and S.~Kim.
\newblock Developing a new hourly forest fire risk index based on catboost in
  south korea.
\newblock \emph{Applied Sciences}, 10\penalty0 (22):\penalty0 8213, 2020.

\bibitem[Kim et~al.(2022)Kim, Choi, Widemann, and Zohdi]{kim2022fast}
Y.~Kim, Y.~Choi, D.~Widemann, and T.~Zohdi.
\newblock A fast and accurate physics-informed neural network reduced order
  model with shallow masked autoencoder.
\newblock \emph{Journal of Computational Physics}, 451:\penalty0 110841, 2022.

\bibitem[LeCun et~al.(2015)LeCun, Bengio, and Hinton]{lecun2015deep}
Y.~LeCun, Y.~Bengio, and G.~Hinton.
\newblock Deep learning.
\newblock \emph{nature}, 521\penalty0 (7553):\penalty0 436--444, 2015.

\bibitem[Mansoor et~al.(2022)Mansoor, Farooq, Kachroo, Mahmoud, Fawzy, Popescu,
  Alyemeni, Sonne, Rinklebe, and Ahmad]{mansoor2022elevation}
S.~Mansoor, I.~Farooq, M.~M. Kachroo, A.~E.~D. Mahmoud, M.~Fawzy, S.~M.
  Popescu, M.~Alyemeni, C.~Sonne, J.~Rinklebe, and P.~Ahmad.
\newblock Elevation in wildfire frequencies with respect to the climate change.
\newblock \emph{Journal of environmental management}, 301:\penalty0 113769,
  2022.

\bibitem[McCarthy et~al.(2021)McCarthy, Tohidi, Aziz, Dennie, Valero, and
  Hu]{mccarthy2021deep}
N.~F. McCarthy, A.~Tohidi, Y.~Aziz, M.~Dennie, M.~M. Valero, and N.~Hu.
\newblock A deep learning approach to downscale geostationary satellite imagery
  for decision support in high impact wildfires.
\newblock \emph{Forests}, 12\penalty0 (3):\penalty0 294, 2021.

\bibitem[Menezes et~al.(2022)Menezes, de~Oliveira, Santos, Russo, de~Souza,
  Roque, and Libonati]{menezes2022lightning}
L.~S. Menezes, A.~M. de~Oliveira, F.~L. Santos, A.~Russo, R.~A. de~Souza, F.~O.
  Roque, and R.~Libonati.
\newblock Lightning patterns in the pantanal: Untangling natural and
  anthropogenic-induced wildfires.
\newblock \emph{Science of The Total Environment}, 820:\penalty0 153021, 2022.

\bibitem[Michael et~al.(2021)Michael, Helman, Glickman, Gabay, Brenner, and
  Lensky]{michael2021forecasting}
Y.~Michael, D.~Helman, O.~Glickman, D.~Gabay, S.~Brenner, and I.~M. Lensky.
\newblock Forecasting fire risk with machine learning and dynamic information
  derived from satellite vegetation index time-series.
\newblock \emph{Science of The Total Environment}, 764:\penalty0 142844, 2021.

\bibitem[Milanovi{\'c} et~al.(2020)Milanovi{\'c}, Markovi{\'c}, Pamu{\v{c}}ar,
  Gigovi{\'c}, Kosti{\'c}, and Milanovi{\'c}]{milanovic2020forest}
S.~Milanovi{\'c}, N.~Markovi{\'c}, D.~Pamu{\v{c}}ar, L.~Gigovi{\'c},
  P.~Kosti{\'c}, and S.~D. Milanovi{\'c}.
\newblock Forest fire probability mapping in eastern serbia: Logistic
  regression versus random forest method.
\newblock \emph{Forests}, 12\penalty0 (1):\penalty0 5, 2020.

\bibitem[Mohajane et~al.(2021)Mohajane, Costache, Karimi, Pham, Essahlaoui,
  Nguyen, Laneve, and Oudija]{mohajane2021application}
M.~Mohajane, R.~Costache, F.~Karimi, Q.~B. Pham, A.~Essahlaoui, H.~Nguyen,
  G.~Laneve, and F.~Oudija.
\newblock Application of remote sensing and machine learning algorithms for
  forest fire mapping in a mediterranean area.
\newblock \emph{Ecological Indicators}, 129:\penalty0 107869, 2021.

\bibitem[Muhammad et~al.(2018)Muhammad, Ahmad, and Baik]{muhammad2018early}
K.~Muhammad, J.~Ahmad, and S.~W. Baik.
\newblock Early fire detection using convolutional neural networks during
  surveillance for effective disaster management.
\newblock \emph{Neurocomputing}, 288:\penalty0 30--42, 2018.

\bibitem[Mukherjee et~al.(2022)Mukherjee, Panja, Dey, and
  Crespo]{mukherjee2022intelligent}
A.~Mukherjee, A.~K. Panja, N.~Dey, and R.~G. Crespo.
\newblock An intelligent edge enabled 6g-flying ad-hoc network ecosystem for
  precision agriculture.
\newblock \emph{Expert Systems}, page e13090, 2022.

\bibitem[Nadeem et~al.(2019)Nadeem, Taylor, Woolford, and
  Dean]{nadeem2019mesoscale}
K.~Nadeem, S.~Taylor, D.~G. Woolford, and C.~Dean.
\newblock Mesoscale spatiotemporal predictive models of daily human-and
  lightning-caused wildland fire occurrence in british columbia.
\newblock \emph{International journal of wildland fire}, 29\penalty0
  (1):\penalty0 11--27, 2019.

\bibitem[Pais et~al.(2021)Pais, Miranda, Carrasco, and Shen]{pais2021deep}
C.~Pais, A.~Miranda, J.~Carrasco, and Z.-J.~M. Shen.
\newblock Deep fire topology: Understanding the role of landscape spatial
  patterns in wildfire occurrence using artificial intelligence.
\newblock \emph{Environmental Modelling \& Software}, 143:\penalty0 105122,
  2021.

\bibitem[Pang et~al.(2022)Pang, Wang, Tay, Liu, Tian, and Yuan]{pang2022masked}
Y.~Pang, W.~Wang, F.~E. Tay, W.~Liu, Y.~Tian, and L.~Yuan.
\newblock Masked autoencoders for point cloud self-supervised learning.
\newblock \emph{arXiv preprint arXiv:2203.06604}, 2022.

\bibitem[Park et~al.(2019)Park, Lee, Yun, Kim, and Kim]{park2019dependable}
J.~H. Park, S.~Lee, S.~Yun, H.~Kim, and W.-T. Kim.
\newblock Dependable fire detection system with multifunctional artificial
  intelligence framework.
\newblock \emph{Sensors}, 19\penalty0 (9):\penalty0 2025, 2019.

\bibitem[Pham et~al.(2020)Pham, Jaafari, Avand, Al-Ansari, Dinh~Du, Yen, Phong,
  Nguyen, Le, Mafi-Gholami, et~al.]{pham2020performance}
B.~T. Pham, A.~Jaafari, M.~Avand, N.~Al-Ansari, T.~Dinh~Du, H.~P.~H. Yen, T.~V.
  Phong, D.~H. Nguyen, H.~V. Le, D.~Mafi-Gholami, et~al.
\newblock Performance evaluation of machine learning methods for forest fire
  modeling and prediction.
\newblock \emph{Symmetry}, 12\penalty0 (6):\penalty0 1022, 2020.

\bibitem[Pozo et~al.(2022)Pozo, Galleguillos, Gonz{\'a}lez, V{\'a}squez, and
  Arriagada]{pozo2022assessing}
R.~A. Pozo, M.~Galleguillos, M.~E. Gonz{\'a}lez, F.~V{\'a}squez, and
  R.~Arriagada.
\newblock Assessing the socio-economic and land-cover drivers of wildfire
  activity and its spatiotemporal distribution in south-central chile.
\newblock \emph{Science of the total environment}, 810:\penalty0 152002, 2022.

\bibitem[Radke et~al.(2019)Radke, Hessler, and Ellsworth]{radke2019firecast}
D.~Radke, A.~Hessler, and D.~Ellsworth.
\newblock Firecast: Leveraging deep learning to predict wildfire spread.
\newblock In \emph{IJCAI}, pages 4575--4581, 2019.

\bibitem[Rodrigues et~al.(2014)Rodrigues, de~la Riva, and
  Fotheringham]{rodrigues2014modeling}
M.~Rodrigues, J.~de~la Riva, and S.~Fotheringham.
\newblock Modeling the spatial variation of the explanatory factors of
  human-caused wildfires in spain using geographically weighted logistic
  regression.
\newblock \emph{Applied Geography}, 48:\penalty0 52--63, 2014.

\bibitem[Sahu and Minz(2022)]{sahu2022self}
K.~Sahu and S.~Minz.
\newblock Self-adaptive-deer hunting optimization-based optimal weighted
  features and hybrid classifier for automated disease detection in plant
  leaves.
\newblock \emph{Expert Systems}, 39\penalty0 (7):\penalty0 e12982, 2022.

\bibitem[Salcedo-Sanz et~al.(2022)Salcedo-Sanz, P{\'e}rez-Aracil, Ascenso,
  Del~Ser, Casillas-P{\'e}rez, Kadow, Fister, Barriopedro, Garc{\'\i}a-Herrera,
  Restelli, et~al.]{salcedo2022analysis}
S.~Salcedo-Sanz, J.~P{\'e}rez-Aracil, G.~Ascenso, J.~Del~Ser,
  D.~Casillas-P{\'e}rez, C.~Kadow, D.~Fister, D.~Barriopedro,
  R.~Garc{\'\i}a-Herrera, M.~Restelli, et~al.
\newblock Analysis, characterization, prediction and attribution of extreme
  atmospheric events with machine learning: a review.
\newblock \emph{arXiv preprint arXiv:2207.07580}, 2022.

\bibitem[Senande-Rivera et~al.(2022)Senande-Rivera, Insua-Costa, and
  Miguez-Macho]{senande2022spatial}
M.~Senande-Rivera, D.~Insua-Costa, and G.~Miguez-Macho.
\newblock Spatial and temporal expansion of global wildland fire activity in
  response to climate change.
\newblock \emph{Nature Communications}, 13\penalty0 (1):\penalty0 1208, 2022.

\bibitem[Shi and Touge(2022)]{shi2022characterization}
K.~Shi and Y.~Touge.
\newblock Characterization of global wildfire burned area spatiotemporal
  patterns and underlying climatic causes.
\newblock \emph{Scientific reports}, 12\penalty0 (1):\penalty0 1--17, 2022.

\bibitem[Short and Finney(2022)]{short2022empirical}
K.~C. Short and M.~A. Finney.
\newblock Empirical records of wildfires caused by firearms use in the united
  states.
\newblock \emph{Fire Safety Journal}, page 103622, 2022.

\bibitem[Solomon et~al.(2022)Solomon, Dube, Stone, Yu, Kinnison, Toon, Strahan,
  Rosenlof, Portmann, Davis, et~al.]{solomon2022stratospheric}
S.~Solomon, K.~Dube, K.~Stone, P.~Yu, D.~Kinnison, O.~B. Toon, S.~E. Strahan,
  K.~H. Rosenlof, R.~Portmann, S.~Davis, et~al.
\newblock On the stratospheric chemistry of midlatitude wildfire smoke.
\newblock \emph{Proceedings of the National Academy of Sciences}, 119\penalty0
  (10):\penalty0 e2117325119, 2022.

\bibitem[Tiefenbacher(2012)]{tiefenbacher2012approaches}
J.~Tiefenbacher.
\newblock \emph{Approaches to Managing Disaster: Assessing Hazards, Emergencies
  and Disaster Impacts}.
\newblock BoD--Books on Demand, 2012.

\bibitem[Tonini et~al.(2020)Tonini, D’Andrea, Biondi, Degli~Esposti,
  Trucchia, and Fiorucci]{tonini2020machine}
M.~Tonini, M.~D’Andrea, G.~Biondi, S.~Degli~Esposti, A.~Trucchia, and
  P.~Fiorucci.
\newblock A machine learning-based approach for wildfire susceptibility
  mapping. the case study of the liguria region in italy.
\newblock \emph{Geosciences}, 10\penalty0 (3):\penalty0 105, 2020.

\bibitem[Torre-Tojal et~al.(2019)Torre-Tojal, Lopez-Guede, and
  Grana~Romay]{torre2019estimation}
L.~Torre-Tojal, J.~M. Lopez-Guede, and M.~M. Grana~Romay.
\newblock Estimation of forest biomass from light detection and ranging data by
  using machine learning.
\newblock \emph{Expert Systems}, 36\penalty0 (4):\penalty0 e12399, 2019.

\bibitem[Van~Wagner et~al.(1987)]{van1987development}
C.~Van~Wagner et~al.
\newblock \emph{Development and structure of the Canadian forest fire weather
  index system}, volume~35.
\newblock 1987.

\bibitem[Vi{\'e} et~al.(2009)Vi{\'e}, Hilton-Taylor, and
  Stuart]{vie2009wildlife}
J.-C. Vi{\'e}, C.~Hilton-Taylor, and S.~N. Stuart.
\newblock \emph{Wildlife in a changing world: an analysis of the 2008 IUCN Red
  List of threatened species}.
\newblock IUCN, 2009.

\bibitem[Xiang et~al.(2022)Xiang, Mao, and Yang]{xiang2022impact}
J.~Xiang, H.~Mao, and B.~Yang.
\newblock Impact assessment and mechanism of water conservancy policy on carbon
  emission performance under the background of artificial intelligence.
\newblock \emph{Expert Systems}, page e13190, 2022.

\bibitem[Xu et~al.(2021)Xu, Liu, and Yan]{xu2021temperature}
Z.~Xu, D.~Liu, and L.~Yan.
\newblock Temperature-based fire frequency analysis using machine learning: A
  case of changsha, china.
\newblock \emph{Climate Risk Management}, 31:\penalty0 100276, 2021.

\bibitem[Yuan et~al.(2021)Yuan, Leirvik, and Wild]{yuan2021global}
M.~Yuan, T.~Leirvik, and M.~Wild.
\newblock Global trends in downward surface solar radiation from spatial
  interpolated ground observations during 1961--2019.
\newblock \emph{Journal of Climate}, 34\penalty0 (23):\penalty0 9501--9521,
  2021.

\bibitem[Zhang et~al.(2021)Zhang, Wang, and Liu]{zhang2021deep}
G.~Zhang, M.~Wang, and K.~Liu.
\newblock Deep neural networks for global wildfire susceptibility modelling.
\newblock \emph{Ecological Indicators}, 127:\penalty0 107735, 2021.

\bibitem[Zhang et~al.(2016)Zhang, Xu, Xu, and Guo]{zhang2016deep}
Q.~Zhang, J.~Xu, L.~Xu, and H.~Guo.
\newblock Deep convolutional neural networks for forest fire detection.
\newblock In \emph{2016 International Forum on Management, Education and
  Information Technology Application}, pages 568--575. Atlantis Press, 2016.

\bibitem[Zohdi(2020)]{zohdi2020machine}
T.~Zohdi.
\newblock A machine-learning framework for rapid adaptive digital-twin based
  fire-propagation simulation in complex environments.
\newblock \emph{Computer Methods in Applied Mechanics and Engineering},
  363:\penalty0 112907, 2020.

\bibitem[Zou et~al.(2019)Zou, O’Neill, Larkin, Alvarado, Solomon, Mass, Liu,
  Odman, and Shen]{zou2019machine}
Y.~Zou, S.~M. O’Neill, N.~K. Larkin, E.~C. Alvarado, R.~Solomon, C.~Mass,
  Y.~Liu, M.~T. Odman, and H.~Shen.
\newblock Machine learning-based integration of high-resolution wildfire smoke
  simulations and observations for regional health impact assessment.
\newblock \emph{International journal of environmental research and public
  health}, 16\penalty0 (12):\penalty0 2137, 2019.

\end{thebibliography}

\clearpage
\onecolumn
\appendix
\medskip

\end{document}